\documentclass[10pt,twocolumn,letterpaper]{article}

\usepackage{iccv}
\usepackage{times}
\usepackage{epsfig}
\usepackage{graphicx}
\usepackage{amsmath}
\usepackage{amssymb}

\usepackage[caption=false]{subfig}

\usepackage{pgfplots}
\DeclareUnicodeCharacter{2212}{−}
\usepgfplotslibrary{groupplots,dateplot,colorbrewer}
\usetikzlibrary{patterns,shapes.arrows,positioning}
\pgfplotsset{compat=newest}
\usepackage{pgf-pie}

\usepackage[protrusion=true, expansion=true, final]{microtype}
\usepackage{booktabs}
\usepackage{tabularx}
\usepackage{arydshln}
\newcommand{\ra}[1]{\renewcommand{\arraystretch}{#1}}
\newcolumntype{Y}{>{\centering\arraybackslash}X}
\usepackage{multirow}
\usepackage{gensymb}

\newcommand{\beginappendixa}{%
        \setcounter{table}{0}
        \renewcommand{\thetable}{A-\arabic{table}}%
        \setcounter{figure}{0}
        \renewcommand{\thefigure}{A-\arabic{figure}}%
     }
\newcommand{\beginappendixb}{%
        \setcounter{table}{0}
        \renewcommand{\thetable}{B-\arabic{table}}%
        \setcounter{figure}{0}
        \renewcommand{\thefigure}{B-\arabic{figure}}%
     }
\newcommand{\beginappendixc}{%
        \setcounter{table}{0}
        \renewcommand{\thetable}{C-\arabic{table}}%
        \setcounter{figure}{0}
        \renewcommand{\thefigure}{C-\arabic{figure}}%
     }
\newcommand{\beginappendixd}{%
        \setcounter{table}{0}
        \renewcommand{\thetable}{D-\arabic{table}}%
        \setcounter{figure}{0}
        \renewcommand{\thefigure}{D-\arabic{figure}}%
     }
\newcommand{\beginappendixe}{%
        \setcounter{table}{0}
        \renewcommand{\thetable}{E-\arabic{table}}%
        \setcounter{figure}{0}
        \renewcommand{\thefigure}{E-\arabic{figure}}%
     }
\newcommand{\beginappendixf}{%
        \setcounter{table}{0}
        \renewcommand{\thetable}{F-\arabic{table}}%
        \setcounter{figure}{0}
        \renewcommand{\thefigure}{F-\arabic{figure}}%
     }
\newcommand{\beginappendixg}{%
        \setcounter{table}{0}
        \renewcommand{\thetable}{G-\arabic{table}}%
        \setcounter{figure}{0}
        \renewcommand{\thefigure}{G-\arabic{figure}}%
     }
     
\usepackage[page]{appendix}
\usepackage[breaklinks=true,bookmarks=false]{hyperref}

\DeclareMathOperator*{\softmax}{softmax}
\DeclareMathOperator*{\softplus}{softplus}
\DeclareMathOperator*{\similarity}{sim}
\DeclareMathOperator*{\argmax}{argmax}
\DeclareMathOperator*{\Gumbel}{Gumbel}

\newcommand{\PLH}{{\mkern-2mu\times\mkern-2mu}}
\DeclareMathOperator*{\oneconv}{1\PLH1\, Conv}
\DeclareMathOperator*{\threeconv}{3\PLH3\, Conv}
\DeclareMathOperator*{\batchnorm}{BN}
\DeclareMathOperator*{\relu}{ReLU}

\iccvfinalcopy % *** Uncomment this line for the final submission

\def\iccvPaperID{****} % *** Enter the ICCV Paper ID here

\ificcvfinal\fi

\begin{document}

%%%%%%%%% TITLE
\title{Exploring Relational Context for Multi-Task Dense Prediction}

\author{David Bruggemann, Menelaos Kanakis, Anton Obukhov, Stamatios Georgoulis, Luc Van Gool\\
ETH Zurich\\
{\tt\small \{brdavid, kanakism, obukhova, georgous, vangool\}@vision.ee.ethz.ch}
% \and
}

\maketitle
\ificcvfinal\thispagestyle{empty}\fi

% no figures in first column
\global\csname @topnum\endcsname 0
\global\csname @botnum\endcsname 0

%%%%%%%%% ABSTRACT
\begin{abstract}
The timeline of computer vision research is marked with advances in learning and utilizing efficient contextual representations.
Most of them, however, are targeted at improving model performance on a single downstream task.
We consider a multi-task environment for dense prediction tasks, represented by a common backbone and independent task-specific heads. 
Our goal is to find the most efficient way to refine each task prediction by capturing cross-task contexts dependent on tasks' relations.
We explore various attention-based contexts, such as global and local, in the multi-task setting and analyze their behavior when applied to refine each task independently.
Empirical findings confirm that different source-target task pairs benefit from different context types.
To automate the selection process, we propose an Adaptive Task-Relational Context (ATRC) module, which samples the pool of all available contexts for each task pair using neural architecture search and outputs the optimal configuration for deployment.
Our method achieves state-of-the-art performance on two important multi-task benchmarks, namely NYUD-v2 and PASCAL-Context.
The proposed ATRC has a low computational toll and can be used as a drop-in refinement module for any supervised multi-task architecture.
\end{abstract}

\section{Introduction}

\begin{figure}[t]
\centering
\includegraphics[width=\linewidth]{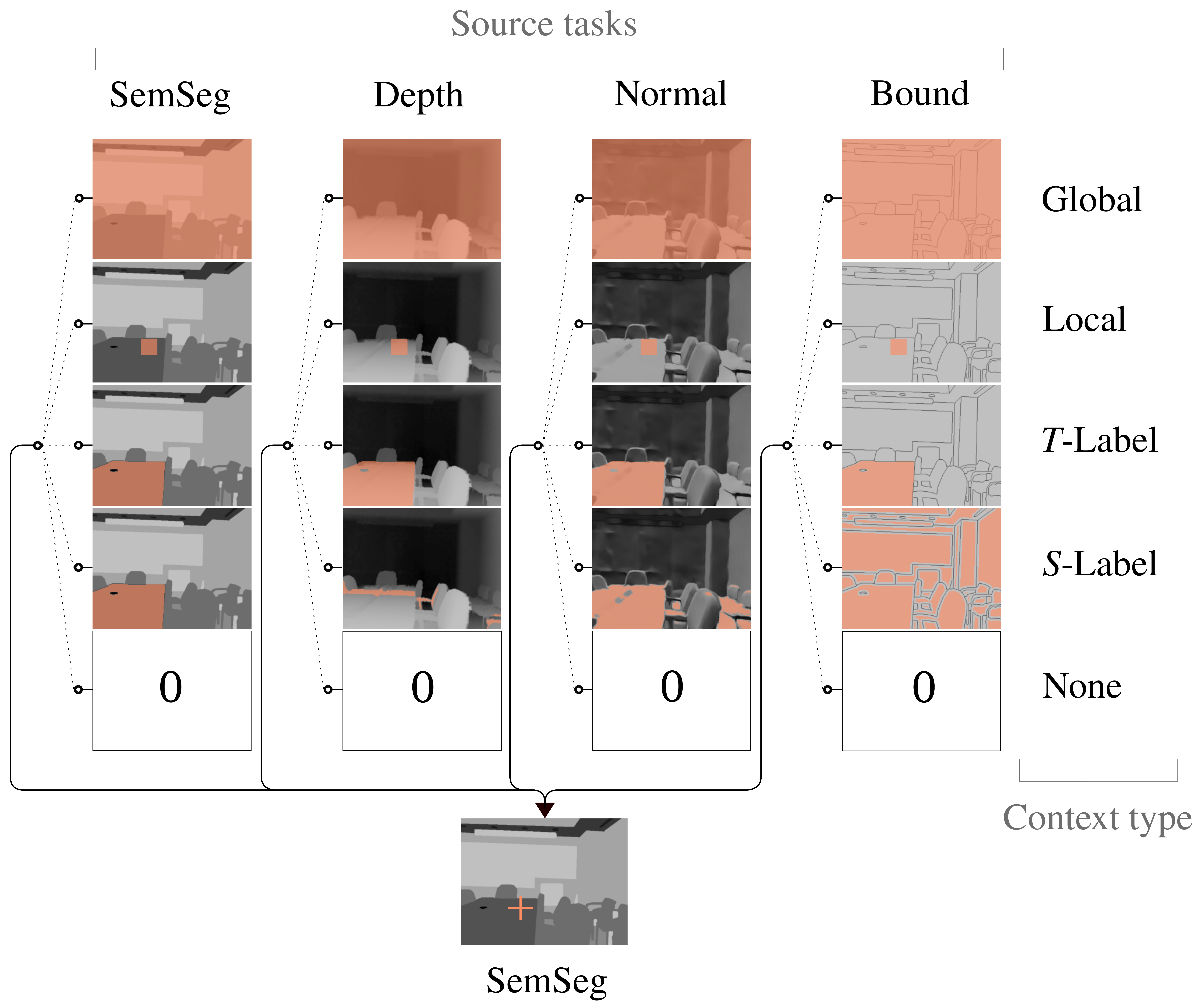}
\caption{Schematic of the task relational context (orange overlay) for the marked pixel (orange cross) of target task semantic segmentation. Our algorithm selects one distillation context type for each source task (dashed lines represent a switch). Alternatively, the connection can be severed by choosing \emph{none}. The procedure is analogous for all other target tasks.}
\label{fig:teaser}
\end{figure}

% In this paper, we study attention-driven contextual representations in the multi-task learning framework. 
The role of context in computer vision is hard to overstate; most notable breakthroughs boil down to a clever extraction~\cite{lowe1999sift}, learning~\cite{lecun1998gradient}, and utilization~\cite{krizhevsky2012imagenet} of contextual representations.
The success of Convolutional Neural Networks (CNN) is largely due to their inherent ability to capture the local context and build very deep~\cite{simonyan2015very} contextual hierarchies within the model.
% Recently, the attention mechanism has been progressively adopted in computer vision tasks~\cite{xu2015show} while steadily replacing the traditional convolutional building blocks~\cite{dosovitskiy2021an}.
Recently, the progressive adoption of the attention mechanism in computer vision~\cite{xu2015show} has brought forth more flexible context descriptions conditioned on the interdependence of individual pixels, while steadily replacing the traditional convolutional building blocks~\cite{dosovitskiy2021an}.

% Multi-Task Learning (MTL)~\cite{Caruana93multitasklearning:} is concerned with reusing representations between tasks.
%
% Several recent works~\cite{xu2018pad,zhang2019pattern,vandenhende2020mti,zhou2020pattern} have shown how to utilize various forms of attention for more efficient context extraction for the downstream tasks. 
%
% These works focus on attention-guided message passing across tasks, task-specific mixtures of self-attention masks, and multi-scale propagation, respectively. 
%
% In this work, however, we are interested in utilizing the modern form of scaled dot-product attention~\cite{vaswani2017attention} and exploring the full range of relations between tasks, which we treat as nodes in a directed graph (Fig.~\ref{fig:teaser}), much like in~\cite{zamir2018taskonomy}. 
%
% We learn a single neural architecture for all tasks with a shared encoder of RGB input, multiple task-specific decoder heads, and our novel Adaptive Relational Context Distillation (ARCD) modules per each task, shown in Fig.~\ref{fig:architecture}. 
%
% ARCD is designed to compute and utilize the single optimal type of relational context out of five possible options for each task with Neural Architecture Search (NAS).

Multi-Task Learning (MTL)~\cite{Caruana93multitasklearning:} is concerned with sharing representations between tasks. Motivated by the observation that representations of visual tasks are often highly correlated~\cite{zhang2019pattern}, recent works~\cite{xu2018pad,vandenhende2020mti} focusing on multi-task dense prediction have extended context extraction across tasks through soft-gated message passing. Referred to as multi-modal distillation in the literature~\cite{xu2018pad}, the idea is to augment the high-level representations of downstream \emph{target tasks} by selectively aggregating complementary features of a set of \emph{source tasks}. The gating function in the distillation thereby learns to focus on useful cross-task information flow. 

Despite their effectiveness, current multi-modal distillation schemes~\cite{xu2018pad,vandenhende2020mti} suffer from two main limitations: (1) The employed gates only regulate information flow based on the source task feature values. As such, the distillation module fails to capture task interactions fully. (2) Each target pixel exclusively receives information from its source counterpart, \ie, the message passing is restricted locally. Compelled by these drawbacks, we propose a new type of attention-driven multi-modal distillation scheme, based on three key contributions:
\begin{enumerate}
    \item Increase the expressivity of the cross-task gate by conditioning it on the interdependence of source and target task pixels. Our multi-modal distillation scheme is therefore \emph{relational}.
    \item Enable global cross-task message passing by enlarging the receptive field of the distillation scheme. We refer to each pixel's distillation receptive field as its \emph{distillation context}.
    \item Customize the distillation context type for each source-target task pair. We formulate five context type candidates (\emph{global}, \emph{local}, \emph{T-label}, \emph{S-label}, \emph{none}) and \emph{adapt} the type automatically with respect to each source-target task pair in a given architecture (see Fig.~\ref{fig:teaser}). 
\end{enumerate}

\begin{figure*}[t]
\centering
\subfloat[]{%
\label{fig:svd_network}{%
\includegraphics[width=0.69\linewidth, trim=0 9.5mm 0 3.25mm, clip]{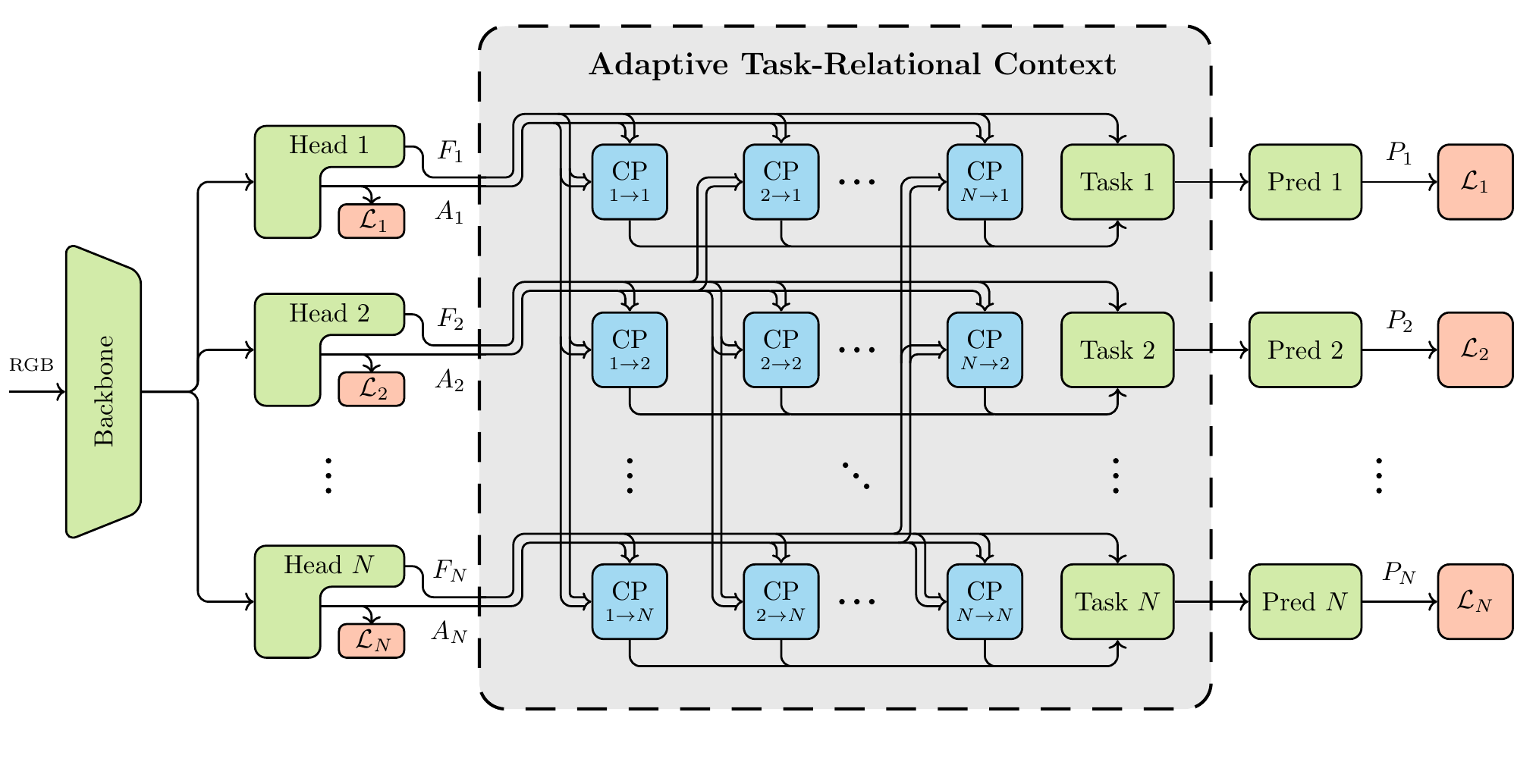}%
}}\hfill%
\subfloat[]{%
\label{fig:tt_network}{%
\includegraphics[width=0.28\linewidth]{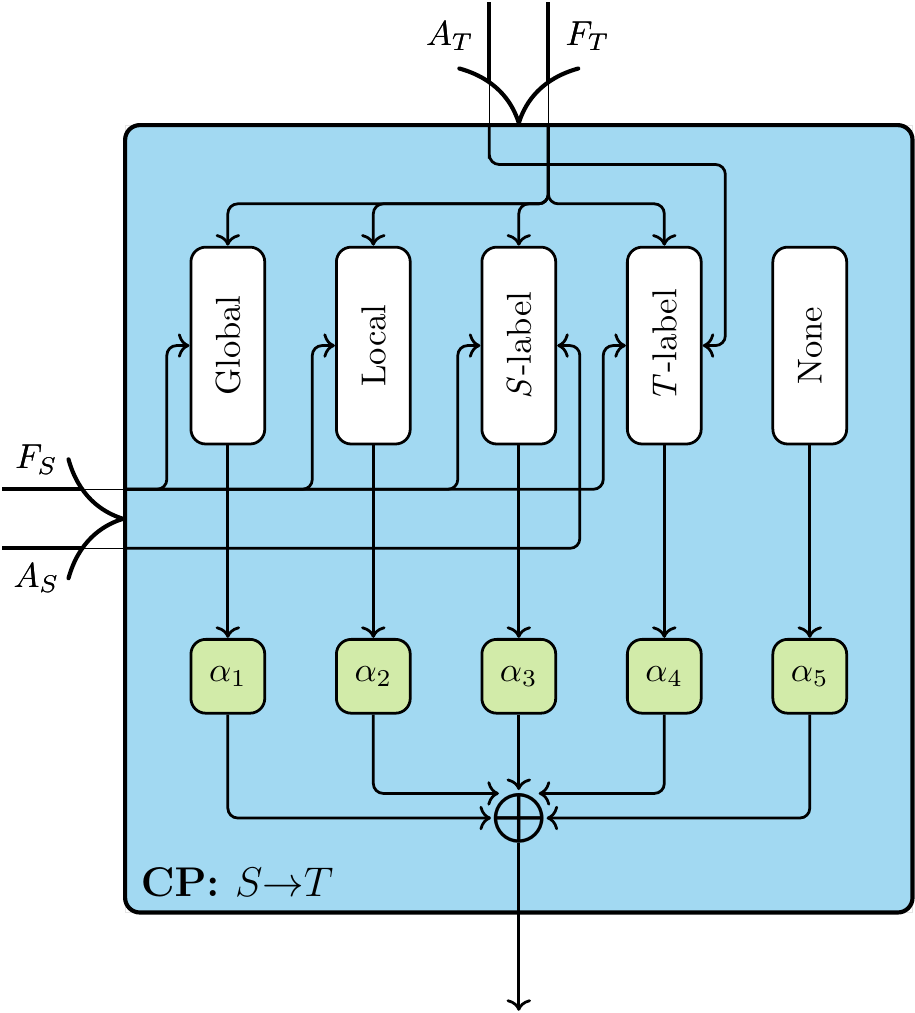}%
}}
\vspace{1ex}
\caption{
\textbf{(a)} Overview of a multi-task network with the proposed Adaptive Task-Relational Context (ATRC) module. The main network can have any topology, provided that the head for each task $n$ produces both the features ($F_n$) for ATRC to refine and the auxiliary prediction ($A_n$). In our experiments we predict $F_n$ and $A_n$ with the main and auxiliary independent heads respectively. Within ATRC each task is routed as target task to $N$ Context Pooling (CP) blocks ($n$-th row of CP blocks) and as source task to $N$ CP blocks ($n$-th column). The outputs of CP blocks are concatenated for each task independently and fed through a projection module (`Task $n$'). The predictions $P_n$ are obtained after processing ATRC outputs with a final layer (`Pred $n$').
\textbf{(b)} Dissection of a CP block, refining target task ($T$) features through source task ($S$) information. During the search stage, the CP block extracts all five contextual representations (white blocks, see Sec.~\ref{subsec:dotproduct}) and returns a convex combination of them. After search convergence, a single context type is sampled via $\argmax$, \ie, the $\alpha_i$ form a one-hot vector.
\textbf{Legend}: 
% Merging single arrows denote the concatenation operation; 
Green blocks denote modules with learned weights, red blocks denote loss functions. Best viewed in color.
%\textcolor{red}{The only thing missing is the reference/explanation of the white blocks in (b).}
}
\label{fig:architecture}
\end{figure*}

Contributions 1 and 2 are addressed by leveraging and adapting the scaled-dot product attention mechanism~\cite{vaswani2017attention} for multi-modal distillation. For contribution 3, we repurpose modern Neural Architecture Search (NAS) methods to automatically find the optimal context type for each source-target task connection. Overall, we present a novel Adaptive Task-Relational Context (ATRC) module which can be used as a drop-in module for CNNs to refine any dictionary of supervised dense prediction tasks. We show its effectiveness empirically with the architecture shown in Fig.~\ref{fig:architecture}: a single neural network for all tasks with a shared backbone of RGB input, multiple task-specific heads, and ATRC distillation modules to refine each task's predictions.

The paper is structured as follows: 
Sec.~\ref{sec:related} provides an overview of related work; 
Sec.~\ref{subsec:design} introduces the architecture of ATRC;
Sec.~\ref{subsec:dotproduct} explains the types of relational contexts in consideration;
Sec.~\ref{subsec:auto} covers the adaptation of the context type through NAS techniques;
Sec.~\ref{sec:experiments} provides the empirical study details and verifies the proposed method state-of-the-art performance on several important benchmarks;
Sec.~\ref{sec:conclusion} concludes the paper.

%% METHOD FIGURE

\section{Related Work}
\label{sec:related}

% \noindent
\textbf{Multi-Task Learning} (MTL) methods employ two main paradigms to learn shared representations: \emph{hard parameter sharing} and \emph{soft parameter sharing}.
%that enable the interaction amongst task-specific representations through a feature fusing mechanism, inducing knowledge transfer between the different tasks \cite{misra2016cross}. 
% Prior art on multi-task learning for dense prediction tasks employs two main paradigms to learn reusable representations: 
% (1) soft parameter sharing, characterized by assigning subsets or linear combinations of weights to various task-specific branches of the overall model architecture, and 
% (2) distillation, which effectively refines the initial predictions in several iterations.
Hard parameter sharing characterizes architectures which typically share the first hidden representations among the tasks while branching to independent task-specific representations at a later stage. 
%Most approaches utilize a shared backbone and split to task-specific heads at a single branch point \cite{kokkinos2017ubernet,neven2017fast,kendall2018multi,chen2018gradnorm,sener2018multi}. 
Most approaches split to task-specific heads at a single branch point \cite{kokkinos2017ubernet,kendall2018multi,chen2018gradnorm,sener2018multi}. 
However, such naive branching can be sub-optimal, raising interest in mechanisms that allow for finely branched architectures \cite{lu2017fully,vandenhende2019branched,bruggemann2020automated}. 
Our work is complementary to these hard parameters sharing methods, since we introduce a module which refines task-specific features.
Soft parameter sharing, in contrast, marks architectures which induce knowledge transfer between separate task-specific networks through feature fusing mechanisms. 
% Soft parameter sharing methods maintain separate task-specific models, which are aim to enhance task-specific representations by fusing features across tasks. 
% Feature fusing can be introduced along the entire network depth \cite{misra2016cross,gao2019nddr,liu2019end}; however, the computational overhead of this can scale poorly as the number of tasks or network depth increases. 
Feature fusing can be introduced along the entire network depth \cite{misra2016cross,gao2019nddr,liu2019end}, whereby computational cost is often a limiting factor. % ; however, the computational overhead of this can scale poorly as the number of tasks or network depth increases. 
Our proposed module can be interpreted as a sophisticated feature fusing mechanism, applied only at a single stage to refine high-level representations.

% More similar to our work, feature fusing modules can also be appended to existing architectures as a refinement stage of high-level task-specific representations.  % \cite{xu2018pad,zhang2019pattern,vandenhende2020mti}.
%This can also be seen as a multi-task decoder.

Several recent MTL works follow a similar strategy:
PAP~\cite{zhang2019pattern} and PSD~\cite{zhou2020pattern} refine task-specific feature maps through global and local self-attention respectively. The employed attention masks are first refined by propagating affinity patterns across tasks and then applied iteratively on the target task feature maps.
%TRL~\cite{zhang2018joint} uses a different self-attention scheme in a framework where two tasks are alternatively processed and refined. 
In contrast to~\cite{zhang2019pattern,zhou2020pattern}, our approach directly attends to source task features by explicitly modeling pairwise interactions between source and target tasks. 
%
%Our cross-task feature interaction, instead, uses multi-model distillation to attend directly to source task features.
More closely related to our work, PAD-Net~\cite{xu2018pad} uses multi-modal distillation to enhance task-specific predictions. Information flow from each source to target task is regulated with a sigmoid-activated gate function.
%Joint Task-Recursive Learning (JTRL)~\cite{zhang2018joint} iteratively refines the predictions of two chained tasks. 
MTI-Net~\cite{vandenhende2020mti} combines the multi-modal distillation module of PAD-Net with a multi-scale refinement scheme to facilitate cross-task talk at multiple scales. However, the gates used in the distillation module of~\cite{xu2018pad,vandenhende2020mti} are functions of the source task features only and operate per pixel. Our method, on the other hand, leverages pairwise task similarities to create more expressive gates through the attention mechanism, while also enabling global cross-task message passing.

\textbf{Attention} was originally developed to improve sentence alignment in neural machine translation~\cite{bahdanau2014neural}.
In computer vision, variants of scaled dot-product attention~\cite{vaswani2017attention} in particular have been used to capture global relationships over the entire pixel space~\cite{wang2018non,bello2019attention,yin2020disentangled}, locally~\cite{ramachandran2019stand}, and even channel-wise~\cite{fu2019dual}. In these approaches, the representation of each target pixel is augmented by aggregating the representations of pixels within the specified context. Each context pixel thereby contributes according to its relation to the target, hence the term \emph{relational context}.
% (as opposed to, \eg, multi-scale context \cite{he2015spatial,chen2017deeplab}). 
Relevant to our work, $A^2$-Net~\cite{chen20182}, ACFNet~\cite{zhang2019acfnet}, and OCR-Net~\cite{yuan2019object} define their own relational context types by grouping pixels into distinct regions (\eg, object class) and attending to prototypical representations of those regions instead.
All of the above mentioned methods focus on attention for a single downstream task and utilize fixed context descriptions. Our work extends these concepts to a multi-task scenario while choosing the optimal relational context type from a pool of candidates for each source-target task pair.

\textbf{Neural Architecture Search} (NAS) automates the process of engineering problem-specific neural network architectures, with the goal of minimizing hand-crafted network design. 
To this end, seminal works use either reinforcement learning~\cite{zoph2016neural,zoph2018learning} or evolutionary~\cite{real2017large,real2019regularized} algorithms to sample promising candidate architectures from a large search space. 
%To this end, \cite{zoph2016neural,zoph2018learning,pham2018efficient} use a recurrent neural network as a controller to sample candidate architectures from a large search space. Through reinforcement learning, the controller is trained to select architectures with maximum expected performance on validation data. Alternatively,~\cite{real2017large,real2019regularized} use evolutionary-based algorithms instead of a controller to mutate promising architectures within the search space gradually. 
Although effective, architecture search with these methods can be very compute-intensive, prompting researchers to explore differentiable NAS~\cite{liu2018darts,xie2018snas,hu2020dsnas}. Instead of a single operation, differentiable NAS uses a convex combination of several operations at a given layer, enabling gradient-based optimization of the search space by training the operation mixing weights. 
The primary contribution of our work is a novel multi-modal distillation module; we thus utilize existing advances in differentiable NAS~\cite{xie2018snas} and a custom search space to automate the context selection for different source-target task pairs.

%%%% I MOVED THE METHOD FIGURE TO RELATED WORK BECAUSE I WANT TO HAVE IT ON PAGE 3

\section{Adaptive Task-Relational Context}
\label{sec:method}

In this section, we describe the proposed Adaptive Task-Relational Context (ATRC) module within a general multi-task learning framework. First, we briefly outline the overall architecture, before dissecting the building blocks of the ATRC module. Finally, we discuss the employed adaptive context type search scheme.

\subsection{Architecture Design}
\label{subsec:design}

% Before we describe the proposed Adaptive Task Relational Context (ATRC) module, we briefly outline the overall architecture, depicted in Fig.~\ref{fig:architecture} c). 
Our ATRC module can be incorporated as a refinement stage in any multi-task neural network (\eg, across multiple scales). 
For transparency we intentionally keep the example configuration simple (see Fig.~\ref{fig:svd_network}):
% While our ATRC module can be incorporated as a refinement stage in any multi-task neural network (\eg, across multiple scales), the example configuration is intentionally kept simple (see Fig.~\ref{fig:svd_network}): 
The backbone is shared among all tasks; shallow heads are used per task to generate task-specific features $F_n$ and auxiliary predictions $A_n$,
%, representing the target and source task feature maps $F_n$, 
where $n \in \left\{1,...,N\right\}$ indexes the task.
In our basic design, we predict $F_n$ and $A_n$ independently, using a $\threeconv$-$\batchnorm$-$\relu$ and $\oneconv$-$\batchnorm$-$\relu$-$\oneconv$ layer respectively.
The role of the $A_n$ is further explained in Sec.~\ref{subsec:similarity_context}.

% Each $F_n$ enters the ATRC module, which complements its target task features with multi-modal information in a residual manner. At last, final predictions are obtained through a classification or regression layer.

% The layout of the ATRC module is depicted in Fig.~\ref{fig:architecture}~(b).
The ATRC module refines the features $F_T$ of each target task $T$ by attending to the features $F_n$ of every available task $n \in \{1,...,N\}$ within a separate Context Pooling (CP) block for each source-target task pair.
Each row of the cartesian grid of CP blocks in Fig.~\ref{fig:svd_network} thus serves to refine one target task $T$, using information from a different source task $S$ in each column.
The self-attention performed in the CP blocks on the diagonal enables the distillation module to additionally capture intra-task relationships.
% The self-attention performed in one of the $N$ CP blocks assigned to $T$ enables the distillation module to additionally capture intra-task relationships.
%
The outputs of all CP blocks within a row are concatenated along the channel dimension, fused with a $\oneconv$-$\batchnorm$ layer, concatenated with the original target task features $F_T$, and processed with $\oneconv$-$\batchnorm$-$\relu$.
Lastly, the refined features are fed through a $\oneconv$ layer (`Pred $T$' in Fig.~\ref{fig:svd_network}) to obtain the final predictions $P_T$.
%Each CP block thus receives the features of the target task, $F_T$, and a source task's features $F_S$.
%In the ATRC module, each $F_n$ is refined using $N$ Context Pooling (CP) blocks (a row of the ATRC matrix in Fig.~\ref{fig:architecture} (a)), acting as $N$ attention heads. 
%
% It receives as inputs the features $x_T$ of the target task $T$ and the features $x_{S_n}$ of each source task $S_n, n \in \left\{1,...,N\right\}$. 
%
% The features from each $S_n$ are aggregated independently: $N$ Context Pooling (CP) blocks ($\widehat{=}\: N$ attention heads, see Sec.~\ref{subsec:dotproduct}) each perform an attention mechanism on a distinct $x_{S_n}$-$x_T$ pair. 
%
%One of the $N$ CP blocks performs self-attention, \ie, $S_n=T$ for that block.
% Note that, the set of source tasks $S_i$ also includes the target task $T$ to additionally introduce self-attention in one of the CP blocks. 
%This allows the distillation module to capture both inter- and intra-task relationships.
%
% Finally, the outputs of the $N$ CP blocks are concatenated and transformed by a 1\texttimes1\,Conv-BN layer to form the output of the ATRC module.

% In \S~\ref{subsec:dotproduct} we outline scaled dot-product attention as the basic component of the CP block, before introducing four adaptations for MTL in \S~\ref{subsec:global_context},\ref{subsec:local_context},\ref{subsec:similarity_context}. The adaptations differ in the context description used (\ie the set of source features each target pixel attends to during distillation).

% \subsection{Scaled Dot-Product Attention for MTL}
\subsection{Context Pooling Block}
\label{subsec:dotproduct}

A CP block aims to extract useful features from one source task $S$ to augment one target task $T$.
% We provide a short outline of scaled dot-product attention and describe how it can be applied for multi-modal feature distillation. 
To this end, each CP block performs at its core a version of scaled dot-product attention, the main component of the widely successful Transformer~\cite{vaswani2017attention}.
%Scaled dot-product attention is the main component of the widely successful Transformer architecture~\cite{vaswani2017attention} and at the core of our CP blocks, shown in Fig.~\ref{fig:architecture} (c). 
Accordingly, the target task feature map $F_T$ and the source task feature map $F_{S}$ are first transformed to queries $q$, keys $k$ and values $v$ using $\oneconv$-$\batchnorm$-$\relu$ layers $f_*$.
\begin{equation}
    q=f_q(F_T),\quad k=f_k(F_S),\quad v=f_v(F_S) 
    \label{eq:value}
\end{equation}
Throughout this paper, we assume that tensors are flattened along the spatial dimension (including $q$, $k$, $v$).
% The $q$, $k$, $v$ are flattened along the spatial dimension. 
A matrix of attention weights $\mathcal{A}$ is generated based on the pairwise similarity between $q$ and $k$ features.
CP block outputs $v'$ are attention-weighted combinations of $v$ features ($d_k$ is the channel dimension of $k$).
\begin{equation}
    v' = \underbrace{\softmax \left(\frac{q k^\intercal}{\sqrt{d_k}} \right)}_{\mathcal{A}} v
    \label{eq:attention}
\end{equation}
In the multi-task setting, the attention weights can be interpreted as modeling the likelihood of feature co-occurrence~\cite{zhang2019co} in transformed target ($q$) and source ($k$) task maps.
The contribution of each source task pixel within the context of the target task pixel is then gated according to the estimated co-occurrence likelihood.
Intuitively, co-occurrence might improve the robustness of target task predictions in ambiguous cases, \eg, for $T$ = `semantic segmentation' and $S$ = `depth estimation', the context of a pixel of class `sky' is more likely to consist of many pixels with large depth. 

% Pairwise attention maps are generated between sets of query $q$ and key $k$ features to model their relation. 
% In a multi-task setting, the queries $q$ are a function of the target task features $x_T$, while the keys $k$ (and values $v$) are functions of the source task features $x_{S_i}$.
% Attention map probabilities are obtained by applying $\softmax$ on the dot product between $q$ and $k$ (scaled by the key dimension $d_k$ to stabilize gradient flow). 
% In a multi-task setting, we are interested in enhancing the features of a target task $T$ through features of a set of source tasks $\mathcal{S} = \left\{S_i | i \in \left\{1...M\right\}\right\}$. 
% This cross-task knowledge transfer can be modeled in a dot-product attention paradigm through the use of multiple attention heads, with each head $H_i$ aggregating features from one source task $S_i$. 
% Specifically, each attention head $H_i$ conducts a separate attention operation (Equation~\ref{eq:attention}) by creating its own queries, keys, and values. The queries $q$ are thereby a function of the target task features $x_T$, while the keys $k$ and values $v$ are both functions of the source task features $x_{S_i}$.
% The outputs of the multiple attention heads are concatenated and merged in a subsequent neural network layer. 
The attention maps in Eq.~\ref{eq:attention} model pixel interactions globally (`all-to-all'), \ie, the distillation context of each pixel is unconstrained. 
Depending on the present source and target task combination, this might not be ideal. 
We therefore introduce four variants of the above attention mechanism in Sec.~\ref{subsec:global_context},~\ref{subsec:local_context},~\ref{subsec:similarity_context}, each characterized by a different context definition. 
%Sec.~\ref{subsec:global_context},~\ref{subsec:local_context},~\ref{subsec:similarity_context}, we introduce the four adaptations of this attention mechanism used in the CP blocks, each one characterized by a different context description (\ie different set of source pixels each target pixel attends to during distillation).
In Sec.~\ref{subsec:auto} we describe how we adapt the CP block for different source-target task pairs.
% Ultimately, a CP block is capable of combining these different context types (\ie, attention mechanisms) to customize task interactions for its source-target task pair (see Sec.~\ref{subsec:auto} and Fig.~\ref{fig:tt_network}).
%
% We use an automated context type selection scheme and is adaptive with respect to the attention mechanism is customized with respect to context type (see Sec.~\ref{subsec:auto} and Fig.~\ref{fig:tt_network}). 
% Ultimately, every CP block attention mechanism is customized with respect to context type (see Sec.~\ref{subsec:auto} and Fig.~\ref{fig:tt_network}). 
% Ultimately, the context type (\ie, attention mechanism) of every CP block is chosen given the target-source task pair. 
%This is achieved using the NAS techniques discussed in Sec.~\ref{subsec:auto}.

\subsubsection{Global Context}
\label{subsec:global_context}
In this case, the distillation context of a specific target pixel is simply every pixel of the source task. Naive implementations of this approach lead to a prohibitively large memory footprint, as the complexity of computing the attention weights scales with $\mathcal{O}(L^2)$, where $L$ is the number of pixels.

To circumvent this issue, we utilize a linearization scheme similar to~\cite{katharopoulos2020transformers}. In particular, we can calculate the attention map for a target pixel $i$ using an arbitrary similarity function $\similarity(\cdot)$ with positive domain instead of $\softmax$.
\begin{equation}
    v'_{i} = \frac{\sum_{j=1}^L\similarity\left(q_{i}, k_{j}\right)v_j}{\sum_{j=1}^L\similarity\left(q_{i}, k_{j}\right)}
\end{equation}
% The $\softmax$ is recoverable by setting $\similarity(u, v) = \exp(u v^T)$. 
This includes all kernel functions $\similarity(q_i,k_j)=\phi(q_i)\phi(k_j)$, which allows us to shift the multiplication order: $\phi(k_j)$ and $v_j$ can be multiplied first and reused for every $\phi(q_i)$, which reduces the overall complexity to $\mathcal{O}(L)$.
% \begin{equation}
%     v'_{i} = \frac{\phi(q_{i})\sum_{j=1}^N\phi(k_{j})^\intercal v_j}{\phi(q_{i})\sum_{j=1}^N \phi(k_{j})^\intercal}
% \end{equation}
In this work, we simply choose a linear kernel $\phi(x) = x$, corresponding to cosine similarity. To avoid numerical issues, we replace the $\relu$ activation functions in $f_q$ and $f_k$ of Eq.~\ref{eq:value} with the smooth approximation $\softplus(x) = \log{\left(1+\exp{(x)}\right)}$.
% Following \cite{katharopoulos2020transformers}, we choose a feature map based on the exponential linear module activation function.
% \begin{equation}
%     \phi(x) = \elu(x) + 1
% \end{equation}

\subsubsection{Local Context}
\label{subsec:local_context}

We can constrain the context to encompass only source pixels spatially close to the target pixel~\cite{ramachandran2019stand}, mimicking the receptive field of a convolution. With $\mathcal{N}_b(i)$ denoting the 2D spatial neighborhood of target pixel $i$ with extent $b$ (we use $b=9\times 9$), the attention formula analogously to Eq.~\ref{eq:attention} is:
\begin{equation}
    %v'_i = \sum_{j \in \mathcal{N}_m(i)} \underbrace{\softmax_{\mathcal{N}_m(i)} \left(\frac{q_i k_j^\intercal}{\sqrt{d_k}} \right)}_{F_j(i)} v_j
    v'_i = \sum_{j \in \mathcal{N}_b(i)} \softmax_{\mathcal{N}_b(i)} \left(\frac{q_i k_j^\intercal}{\sqrt{d_k}} \right) v_j
\end{equation}
This operation resembles a convolution with a spatially-adaptive filter~\cite{su2019pixel}---the attention map.

\subsubsection{Label Context}
\label{subsec:similarity_context}

Both the global and local relational contexts are spatially defined, \ie, distillation is conducted through a spatial attention mask. \emph{Label} context, on the other hand, is defined in label space, meaning that we (1) partition the label space into a set of disjoint label regions, (2) find a prototypical representation for each region, and (3) relate each pixel to each region prototype. This concept has been applied to semantic segmentation in~\cite{zhang2019acfnet,yuan2019object}. In this section we generalize it to any dense prediction task and explore its potential for MTL.

Partitioning the label space is straightforward for classification tasks, \ie, the label regions can be equivalent to the classes. For regression tasks, however, we need to discretize the continuous label space. Consequently, we bin the values on a logarithmic scale for depth prediction and cluster predictions on the unit sphere using k-means for surface normal estimation (see Appendix~\ref{sec:regression_tasks} for details).

We follow the approach of OCR-Net~\cite{yuan2019object} for supervised learning of the region prototypes for each task $n$: Specifically, auxiliary prediction heads calculate the spatial maps $A_n \in \mathbb{R}^{L\times R_n}$ (see Fig.~\ref{fig:svd_network}), where each entry indicates the degree to which a pixel  $l \in \left\{1,...,L\right\}$ belongs to a label region $r \in \left\{1,...,R_n\right\}$. During training, these maps are learned with ground truth supervision using a cross-entropy loss. The resulting maps $A_n$ are normalized using spatial $\softmax$ to obtain $\hat{A}_n$, representing the spatial probability density of each label region $r$.
% The above process is repeated for each task $t$, such that we obtain separate region density maps $\hat{A}_t$. 
In a multi-task setup, we can then choose to define the label regions in either target or source task label space:
% 
% and find a feature representation for each context region within the prediction. This is done in practice by using a dense prediction head which assigns the pixels softly to the context regions. The resulting representations are then aggregated per context region to form `average` context feature representations. Subsequently, these contextual representations will serve as key and value features in a scaled dot-product attention mechanism, which complements task-specific features in a second dense prediction head.
% 
% In a multi-task setting, \emph{similarity context} can be employed in two ways. In the following, assume for illustration purposes that we augment the target task semantic segmentation (\emph{Semseg}) using contextual representations of the source task depth prediction (\emph{Depth}):

\textbf{$T$-label}. In this approach, label regions are defined in target task ($T$) space. Source task features are spatially aggregated using target task spatial maps $\hat{A}_T$, yielding the region prototypes $p_S \in \mathbb{R}^{R_T\times C}$, where $C$ is the source task channel dimension. %Soft-weighted aggregation of source task features $x_{S_i}$ yields the prototype $x'_{S_i}[r]$ of region $r$ within target task label space.
\begin{equation}
    p_{S} = \hat{A}_T^\intercal F_{S}
    \label{eq:labelc}
\end{equation}
$p_{S}$ is then substituted for $F_{S}$ in Eq.~\ref{eq:value} to obtain $k$ and $v$. 
% Assuming a scenario where $T$ is semantic segmentation and $S_i$ is depth prediction, this would enhance each target pixel's representation based on an aggregated \emph{Depth} representation of all pixels belonging to the target pixel semantic class (\eg, the average depth feature of all `car` pixels).
% we define the context using label similarity on the target task, \emph{Semseg}, thus the context would consist of all pixels of the same semantic class as the target pixel.

\textbf{$S$-label.} Alternatively, source task features can also be aggregated via source task ($S$) spatial maps $\hat{A}_{S}$, by substituting $\hat{A}_{S}$ for $\hat{A}_T$ in Eq.~\ref{eq:labelc}.
% \begin{equation}
%     x'_{S_i}[r] = M_r^{(S_i)} x_{S_i}
% \end{equation}
% Again assuming $T$ is semantic segmentation and $S_i$ is depth prediction, in this case we aggregate all $Depth$ feature representations of the pixels with similar depth as the target pixel.

% The context is defined using label similarity on the source task \emph{Depth}. Contextual representations are therefore built independently of the target task \emph{Semseg}.

The key difference between the two approaches is best illustrated with an example: 
Assuming target task semantic segmentation and source task depth estimation, the $T$-label context groups depth features according to object class and makes each target pixel attend to the prototypical depth features for each object class (\eg, the representative depth feature of all `car' pixels). Conversely, the $S$-label context simply groups depth features according to their depth, enabling semantic features to interact with entire depth regions.

% enabling semantic features to interact given knowledge of their depth (\eg, differentiating between objects located at a similar depth compared to others placed further back).

% Assuming target task depth estimation and source task semantic segmentation, the $T$-label context groups semantic features according to their depth and makes each target pixel attend to the prototypical object class for each depth range (\eg, 

\begin{figure}[t]
\centering
\includegraphics[width=\linewidth]{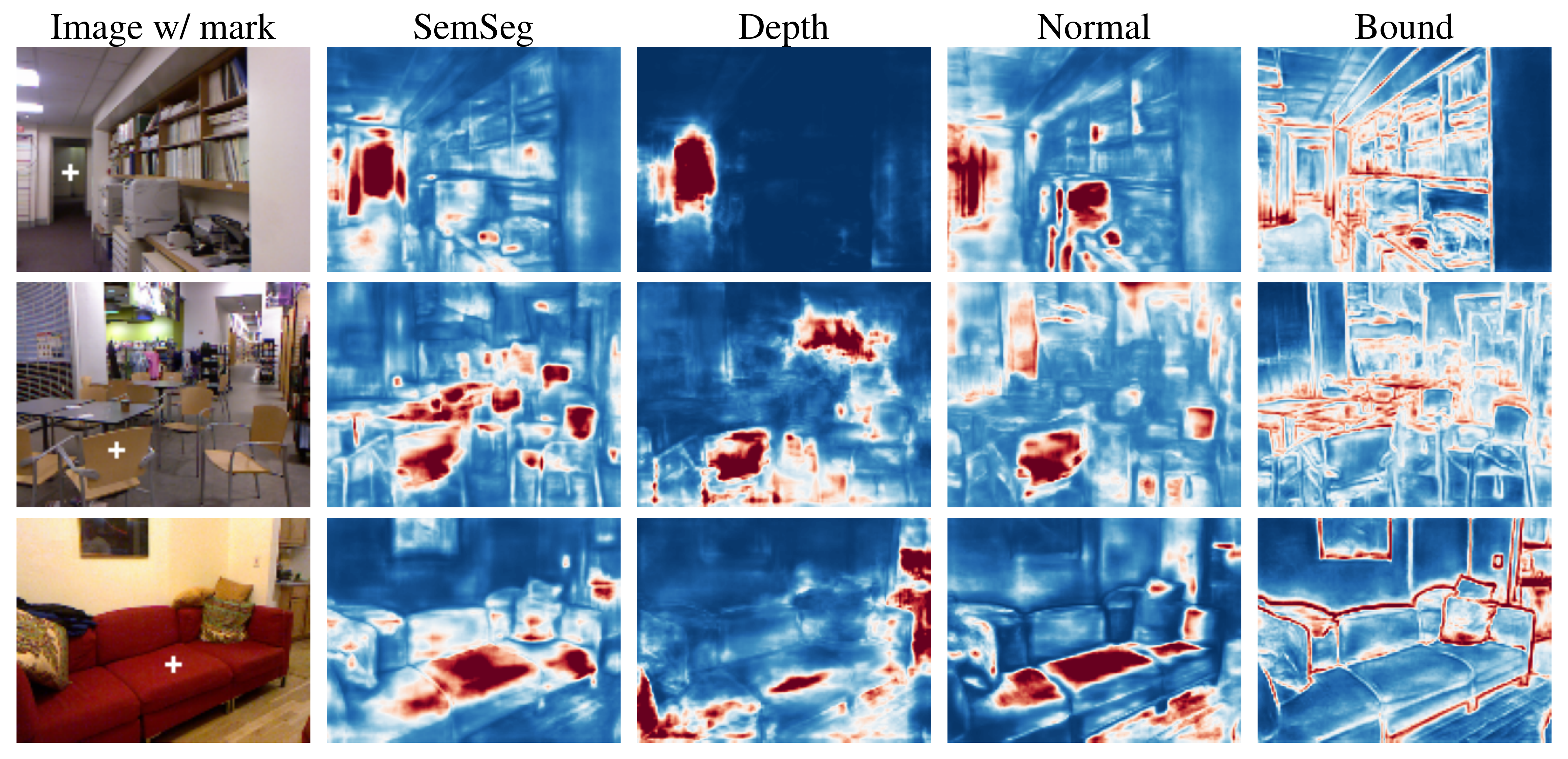}
   \caption{Heatmaps showing label context attention maps relating to the pixel marked with a white cross in the left image, \ie we visualize the corresponding row of $\mathcal{A}$ in Eq.~\ref{eq:attention}. For each target task we visualize the self-attention maps only.}
\label{fig:visual}
\end{figure}

We visualize example self-attention maps for a single target pixel (white cross) of a trained label context distillation model in Fig.~\ref{fig:visual}. The maps illustrate that the model learns to focus on context pixels within distinct label regions.

\subsection{Automated Context Type Selection}
\label{subsec:auto}

While all presented context types could help improve target task features, some might be more effective than others in specific scenarios. 
Therefore, CP blocks are designed to tailor their context type (attention mechanism) to the present source-target task pair.
%
% In principle, such customization could for example be achieved through a mixture-of-experts~\cite{jacobs1991adaptive} or negative correlation learning~\cite{liu1999ensemble} setup.
% %
% Instead, we opt for differentiable NAS techniques to automatically select a single context type for each CP block, by optimizing a supergraph encompassing all options (see Fig.~\ref{fig:tt_network}).
In this paper we opt for differentiable NAS techniques to automatically select a single context type for each CP block, by optimizing a supergraph encompassing all options (see Fig.~\ref{fig:tt_network}). However, a CP block is not limited to a single context type \emph{per se} and could instead refine predictions given a combination of context types in a static~\cite{liu1999ensemble, xie2017aggregated} or even dynamic~\cite{jacobs1991adaptive} fashion.
%
% Instead of manually selecting context types, we exploit differentiable NAS techniques to automatically find the best context for each task pair by optimizing a supergraph encompassing all options. 

Our search space consists of five candidates in each CP block: \emph{global}, \emph{local}, \emph{T-label}, \emph{S-label}, and a \emph{none} operation. The \emph{none} operation simply severs the information flow between two tasks, which can prevent task interference, a common problem in MTL~\cite{kokkinos2017ubernet,kanakis2020reparameterizing}. Operation selection in a CP block $j$ can be formulated as a multiplication of all candidates $O_{j}$ with a one-hot vector $Z_{j}$ sampled from the categorical distribution $p_{\alpha_{j}}(Z_{j})$.
\begin{equation}
    \tilde{O}_{j} = Z_{j}^\intercal O_{j}
\end{equation}
Continuous relaxation of the search space (while maintaining this sampling process) is achieved through the Gumbel-Softmax gradient estimator~\cite{maddison2016concrete,jang2016categorical}, yielding a softened one-hot random variable $\hat{Z}_{j}$.
\begin{equation}
    \hat{Z}_{j}^{(i)} = \frac{\exp{\left(\left(\log \alpha_{j}^{(i)} + G_{j}^{(i)}\right)/\lambda\right)}}{\sum_{u=1}^5 \exp{\left(\left(\log \alpha_{j}^{(u)} + G_{j}^{(u)}\right)/\lambda\right)}}
    \label{eq:gumbel}
\end{equation}
$G_{j}^{(u)} \sim \Gumbel(0,1)$ is a Gumbel random variable, and $\lambda$ is the $\softmax$ temperature. In our case, the architecture parameters $\alpha$ are updated in the same round of backpropagation as the network weights (single-level optimization). A more detailed discussion of Gumbel-Softmax for differentiable NAS is provided in~\cite{xie2018snas}.

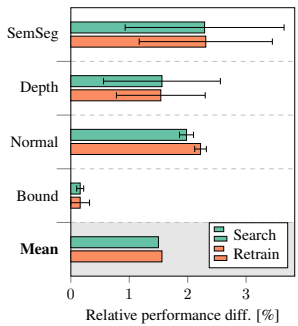
\begin{figure}[t]
    \begin{minipage}[c]{0.48\linewidth}
        \resizebox{\linewidth}{!}{%
            % This file was created by tikzplotlib v0.9.8.
\begin{tikzpicture}

% \definecolor{color0}{rgb}{0.298039215686275,0.447058823529412,0.690196078431373}
% \definecolor{color1}{rgb}{0.333333333333333,0.658823529411765,0.407843137254902}

\definecolor{color0}{rgb}{0.4,0.76078431372549,0.647058823529412}
\definecolor{color1}{rgb}{0.988235294117647,0.552941176470588,0.384313725490196}

\pgfdeclarelayer{background}
\pgfdeclarelayer{foreground}
\pgfsetlayers{background,main,foreground} 

\begin{axis}[
    scale only axis=true,
    height=6cm,
    width=5cm,
    xbar,
    bar width=7pt,
    tick align=outside,
    tick pos=left,
    ytick = {1,2,3,4,5},
    yticklabels={\textbf{Mean},Bound,Normal,Depth,SemSeg},
    ymin=0.5, ymax=5.5,
    y grid style=dashed,
    yminorgrids,
    xlabel={Relative performance diff. [\%]},
    xmin=0, xmax=3.8325,
    major y tick style = {opacity=0},
    minor y tick num = 1,
    minor tick length=2ex,
    reverse legend,
    legend cell align={left},
	legend style={column sep=2pt},
	legend pos=south east
]

\addplot[draw=black,fill=color1,error bars/.cd, x dir=both, x explicit]
	coordinates {
	(1.56,1) % += (0.24,0) -=(0.24,0)};
	(0.16,2)  += (0.16,0) -=(0.16,0)
	(2.22,3)  += (0.10,0) -=(0.10,0)
	(1.54,4)  += (0.76,0) -=(0.76,0)
	(2.31,5) += (1.14,0) -=(1.14,0) 
	};
\addplot[draw=black,fill=color0,error bars/.cd, x dir=both, x explicit] 
	coordinates {
	(1.50,1) % += (0.34,0) -=(0.34,0)};
	(0.16,2) += (0.06,0) -=(0.06,0) 
	(1.98,3) += (0.12,0) -=(0.12,0) 
	(1.56,4) += (1.00,0) -=(1.00,0)
	(2.29,5) += (1.36,0) -=(1.36,0) 
	};
\legend{Retrain,Search}

\begin{pgfonlayer}{background}
\draw [draw=none,fill=black!10,on layer=axis background] (0,0.5) rectangle (3.8325,1.5);
\end{pgfonlayer}

\end{axis}

\end{tikzpicture}
        }%
    \end{minipage}\hfill
    \begin{minipage}[c]{0.49\linewidth}
        \caption{Performance comparison of the models sampled from the supergraph at the end of the context type search \vs after retraining. The chart shows mean and std.\ of the relative performance improvement w.r.t.\ single task (ST) models: $(M_{m}-M_{ST})/M_{ST}$ for model $m$ and `higher = better' metric $M$, and \emph{vice versa} for `lower = better'.}
        \label{fig:agreement}
    \end{minipage}
\end{figure}

Empirically, samples from $\alpha$-distributions trained with Gumbel-Softmax exhibit large variance after convergence, leading to unstable evaluation of sampled subgraphs. We thus use a two-pronged strategy to counteract this problem: (1) Similarly to~\cite{gao2020mtl}, we adopt entropy regularization on $p_{\alpha_{j}}(Z_{j})$ to explicitly control the sampling variance. Instead of the commonly employed candidate operation pretraining, we can simply start the architecture search from scratch with a negative regularization weight to enforce a uniform $\alpha$-distribution. The weight is gradually increased to a positive value during training to ultimately incentivize low-entropy solutions, which imply a low variance as the architecture is sampled from the supergraph. (2) We stop the architecture sampling process in CP block $j$ completely once $p_{\alpha_{j}}$ has reached a low-entropy solution. After a defined threshold is surpassed, we fix the block selection procedure in $j$ using $\argmax$. 
Using this strategy, we obtain high-performing architectures directly during the search stage (see Fig.~\ref{fig:agreement}), demonstrating that our search objective is well defined. Nevertheless, for a fair comparison, we still retrain the discovered architectures from scratch---as is common practice~\cite{liu2018darts,xie2018snas}.

%%% TABLE MOVED HERE BC I WANT IT ON PAGE 6
\begin{table*}
\begin{center}
\resizebox{\linewidth}{!}{%

\ra{1.3}

% \begin{tabular}{@{}lYYcYYcYYcYYcYYcr@{}}\toprule
\begin{tabular}{@{}lcccccccccccccccr@{}}\toprule
\multirow{2}{*}{Distillation module} & \multicolumn{2}{c}{Resource} & \phantom{abc} & \multicolumn{2}{c}{SemSeg~$\uparrow$} & \phantom{abc} & \multicolumn{2}{c}{Depth~$\downarrow$} & \phantom{abc} & \multicolumn{2}{c}{Normal~$\downarrow$} & \phantom{abc} & \multicolumn{2}{c}{Bound~$\uparrow$} & \phantom{abc} & \multirow{2}{*}{$\Delta_m$~[\%]~$\uparrow$} \\
\cmidrule{2-3} \cmidrule{5-6} \cmidrule{8-9} \cmidrule{11-12} \cmidrule{14-15} & Params (M) & MAdds (G) && mean & std. && mean & std. && mean & std. && mean & std.\\ \midrule
None (single task baseline) & 16.09 & 40.93 && 38.02 & 0.14 && 0.6104 & 0.0041 && 20.94 & 0.08 && 76.22 & 0.07 && 0.00 \\
None (multi-task baseline) & \phantom{0}4.52 & 17.59 && 36.35 & 0.26 && 0.6284 & 0.0034 && 21.02 & 0.06 && 76.36 & 0.05 && -1.89 \\ \hdashline\noalign{\vskip 0.5ex}
Cross-Stitch~\cite{misra2016cross} & \phantom{0}4.52 & 17.59 && 36.34 & 0.55 && 0.6290 & 0.0051 && 20.88 & 0.04 && 76.38 & 0.07 && -1.75 \\
PAP~\cite{zhang2019pattern} & \phantom{0}4.54 & 53.04 && 36.72 & 0.31 && 0.6178 & 0.0065 && 20.82 & 0.03 && 76.42 & 0.07 && -0.95 \\
PSD~\cite{zhou2020pattern} & \phantom{0}4.71 & 21.10 && 36.69 & 0.55 && 0.6246 & 0.0036 && 20.87 & 0.07 && 76.42 & 0.13 && -1.30 \\
PAD-Net A~\cite{xu2018pad} / NDDR-CNN~\cite{gao2019nddr} & \phantom{0}4.59 & 18.68 && 36.72 & 0.31 && 0.6288 & 0.0037 && 20.89 & 0.02 && 76.32 & 0.07 && -1.51 \\
PAD-Net B~\cite{xu2018pad} & \phantom{0}5.02 & 25.18 && 36.70 & 0.16 && 0.6264 & 0.0021 && 20.85 & 0.03 && 76.50 & 0.06 && -1.33 \\
PAD-Net C~\cite{xu2018pad} / MTI-Net~\cite{vandenhende2020mti} & \phantom{0}5.50 & 32.42 && 36.61 & 0.15 && 0.6270 & 0.0048 && 20.85 & 0.03 && 76.38 & 0.07 && -1.44 \\ \hdashline\noalign{\vskip 0.5ex}
Global relational context & \phantom{0}4.73 & 21.43 && 38.30 & 0.65 && 0.6007 & 0.0073 && 20.60 & 0.07 && 76.26 & 0.05 && 1.00 \\
Local relational context & \phantom{0}4.73 & 22.19 && 36.79 & 0.29 && 0.6260 & 0.0044 && 20.91 & 0.06 && 76.44 & 0.05 && -1.34 \\
$T$-label relational context & \phantom{0}5.06 & 25.91 && 38.88 & 0.31 && 0.6059 & 0.0014 && 20.48 & 0.05 && 76.30 & 0.06 && 1.33 \\
$S$-label relational context & \phantom{0}5.06 & 25.91 && 38.33 & 0.64 && 0.6006 & 0.0019 && 20.56 & 0.06 && 76.26 & 0.05 && 1.07 \\ \hdashline\noalign{\vskip 0.5ex}
ATRC (ours) & \phantom{0}5.06 & 25.76 && 38.90 & 0.43 && 0.6010 & 0.0046 && 20.48 & 0.02 && 76.34 & 0.12 && \textbf{1.56} \\
\bottomrule
\end{tabular}%
}
\end{center}
\caption{Controlled distillation module comparison on NYUD-v2 with a HRNet18 backbone. For all models except the single task baseline, a shared encoder and small task-specific heads are used (Sec.~\ref{subsec:design}). We insert the different distillation modules before the final prediction layer.}
\label{tab:control}
\end{table*}

\section{Experiments}
\label{sec:experiments}

We briefly review the experimental setup, before presenting empirical studies. Training details are provided in Appendix~\ref{sec:train_details} and reference code is available at \mbox{\url{https://github.com/brdav/atrc}}.

\subsection{Setup}

\textbf{Datasets.} Experiments are conducted on two widely-used dense prediction datasets: (1) \emph{NYUD-v2}~\cite{silberman2012indoor}, which consists of 795 training and 654 testing images of indoor scenes, with annotations for semantic segmentation (`SemSeg'), depth estimation (`Depth'), surface normal estimation (`Normal'), and boundary detection (`Bound'). (2) \emph{PASCAL-Context}~\cite{chen2014detect}, a split of the larger PASCAL dataset~\cite{everingham2010pascal}, providing 4998 training and 5105 testing images, labeled for semantic segmentation, human parts segmentation (`PartSeg'), saliency estimation (`Sal'), surface normal estimation, and boundary detection. We use the distilled saliency and surface normal labels of~\cite{maninis2019attentive}.

\textbf{Backbones.} We test our framework using several backbones: HRNetV2-W18-small (HRNet18), HRNetV2-W48 (HRNet48)~\cite{wang2020deep}, and ResNet-50~\cite{he2016deep}.
%To demonstrate the flexibility and generalizability of ATRC modules, we append them to several backbones: HRNetV2-W18-small (HRNet18), HRNetV2-W48 (HRNet48)~\cite{wang2020deep}, and ResNet-50~\cite{he2016deep}.

\textbf{Metrics.} We evaluate `Semseg' and `PartSeg' with mean intersection over union, `Depth' with root mean square error, `Normal' with mean angular error, `Sal' with maximum F-measure as in~\cite{achanta2009frequency}, and `Bound' with the optimal-dataset-scale F-measure of~\cite{martin2004learning}. All experiments in this paper are repeated five times; the mean is reported for every metric (in Table~\ref{tab:control} also the standard deviation). To quantify overall multi-task performance for $N$ tasks, we adopt the average per-task performance drop ($\Delta_m$) with respect to single task baselines $b$ for model $m$~\cite{maninis2019attentive}: 
$\Delta_m = \frac{1}{N} \sum_{i=1}^N (-1)^{\gamma_i} (M_{m,i}-M_{b,i})/M_{b,i}$.
% \begin{equation}
%     \Delta_m = \frac{1}{\tau} \sum_{i=1}^\tau (-1)^{\gamma_i} (M_{m,i}-M_{b,i})/M_{b,i}
% \end{equation}
$\gamma_i=1$ if lower is better for metric $M_i$ and $\gamma_i=0$ otherwise. 

\subsection{Distillation Module Benchmarking}

In Table~\ref{tab:control} we conduct a series of controlled experiments to assess the effectiveness of different distillation modules fairly. Using a HRNet18 backbone, we alter the MTL architecture design described in Sec.~\ref{subsec:design} only by replacing the ATRC module with other distillation modules. For the baselines, no distillation module is used.

As expected, all investigated distillation modules outperform the trivial multi-task baseline in terms of multi-task performance $\Delta_m$. Furthermore, most relational context modules fare significantly better than their alternatives. Excepting local relational context, augmenting the multi-task network with relational context beats the single task baseline while maintaining a far lower computational footprint.

Table~\ref{tab:control} also reveals that no single relational context type dominates for every task. This suggests that a more fine-grained context customization for each individual source-target task pair could improve overall performance. Indeed, applying our automated context type selection (Sec.~\ref{subsec:auto}), ATRC, produces the best result in multi-task performance.

\begin{figure*}[t]
\centering
\includegraphics[width=\linewidth]{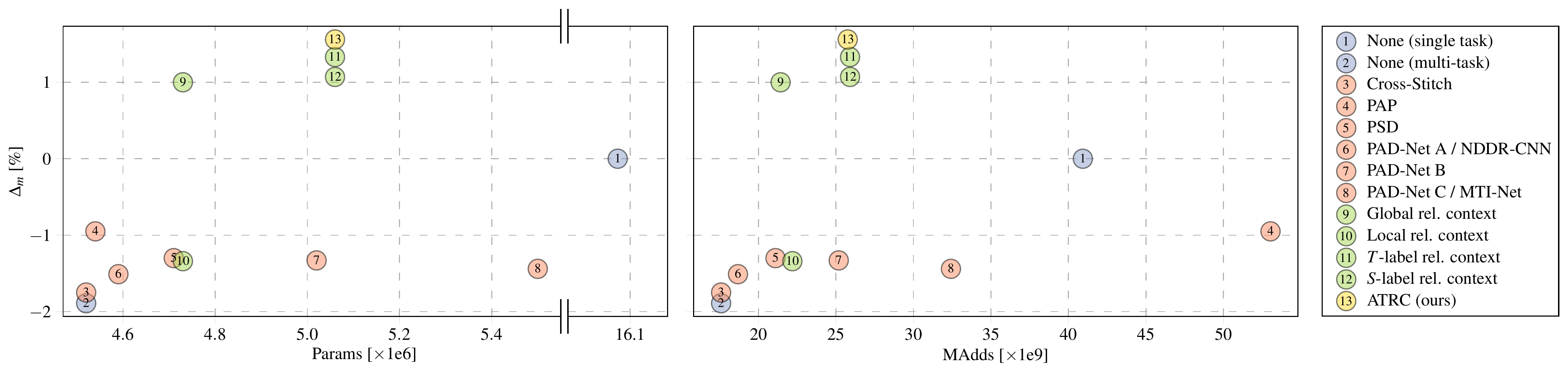}
   \caption{Distillation module resource analysis using an HRNet18 backbone on NYUD-v2. We plot multi-task performance $\Delta_m$ \vs number of parameters (left) and MAdds (right) for multi-task models with different distillation modules inserted before the final prediction layer.}
\label{fig:res}
\end{figure*}

Fig.~\ref{fig:res} visualizes the resource cost of the various distillation modules by plotting the multi-task performance \vs number of parameters and multiply-add operations (MAdds). The computational overhead of the relational context modules---and most other distillation modules---remains low compared to single task networks. Our ATRC combines the benefits of all the relational context modules by maximizing performance while remaining bounded in terms of resource cost.
%, with the exception of the iterative PAP~\cite{zhang2019pattern} module, which surpasses the single-task baseline in terms of MAdds.

% \subsection{Search Algorithm Verification}
% 
% We verify the context search procedure with two checks, presented in Fig.~\ref{fig:agreement}. In the left plot, we show the agreement in context type choice between five search runs on NYUD-v2.
% We quantify agreement using Light's kappa~\cite{light1971measures}, a robust statistic for measuring inter-rater reliability. 
% Following~\cite{light1971measures}, we quantify agreement using the mean of Cohen's kappa~\cite{cohen1960coefficient} for all search run pairs (Light's kappa). Notably, we observe that the agreement rises from 0.48 (`moderate agreement`~\cite{krippendorff2018content}) all the way up to 1.0 as source-target task connections are progressively ignored in order of importance (see Sec.~\ref{subsec:importance} for importance inspection), confirming that the algorithm becomes more selective for important choices. The right plot of Fig.~\ref{fig:agreement} displays a performance comparison of the model directly after the search \vs a retrained model. Owing to the measures taken to minimize post-search sampling variance (Sec.~\ref{subsec:auto}), the search model performs almost on par with the retrained one.

\subsection{Comparison with State-of-the-Art}

To validate the proposed ATRC module, we present experimental comparisons with the following baselines across a number of scenarios: separate single task networks, multi-task network (shared backbone; task-specific heads; no distillation) and the state-of-the-art MTI-Net~\cite{vandenhende2020mti}.
% in this section experimental comparisons with single and multi-task baselines in addition to the state-of-the-art MTI-Net~\cite{vandenhende2020mti}, across a number of scenarios.
Tables~\ref{tab:nyud18} and~\ref{tab:nyud48} display the results obtained on the NYUD-v2 dataset, using HRNet18 and HRNet48 backbones respectively, while Table~\ref{tab:pascal18} shows PASCAL-Context results using HRNet18. MTI-Net uses a large-scale decoder head consisting of two separate stages: the Feature Propagation Module (FPM) and a multi-scale multi-modal distillation module (the analog of our ATRC module). To ensure a fair comparison, we apply our method on both the basic architecture described in Sec.~\ref{subsec:design}, as well as the backbone complemented with the FPM (+174\% and +79\% in number of parameters for HRNet18 and HRNet48 respectively). 

In all investigated cases, ATRC enhances performance significantly compared to the multi-task baseline. 
Furthermore, our method combined with the FPM consistently outperforms MTI-Net, even though MTI-Net applies multi-modal distillation on four scales, while we only distill on the largest scale (causing our model to be more parameter efficient, \eg, \mbox{-22\%} in Table~\ref{tab:nyud18}). 
This implies that task interactions can be adequately captured at a single scale for distillation, provided that the backbone is able to extract and fuse multi-scale information effectively (like HRNet).
% This demonstrates that multi-scale distillation is not required to attain state-of-the-art performance, at least when refining features extracted with a backbone using multi-scale fusion (as HRNet) and a FPM, which already mixes task information at multiple scales.

Overall, the multi-task approaches are less effective compared to single task baselines on the PASCAL-Context dataset. This finding is in agreement with other works~\cite{maninis2019attentive,vandenhende2020mti} and could be attributed to the larger and more diverse task dictionary. Nevertheless, the ranking order of the multi-task approaches in terms of multi-task performance remains consistent with the results obtained for NYUD-v2.

\begin{table}
\begin{center}
\resizebox{\linewidth}{!}{%

\ra{1.3}
\large

% \begin{tabular}{@{}lccrrcrrcrrcrrcr@{}}\toprule
% Model & FPM & \phantom{} & \multicolumn{2}{c}{SemSeg~$\uparrow$} & \phantom{a} & \multicolumn{2}{c}{Depth~$\downarrow$} & \phantom{a} & \multicolumn{2}{c}{Normal~$\downarrow$} & \phantom{a} & \multicolumn{2}{c}{Boundary~$\uparrow$} & \phantom{a} & $\Delta_m$~[\%]~$\uparrow$ \\
% \cmidrule{4-5} \cmidrule{7-8} \cmidrule{10-11} \cmidrule{13-14} &&& mean & std. && mean & std. && mean & std. && mean & std.\\ \midrule
% Single task &&& 38.02 & 0.14 && 0.6104 & 0.0041 && 20.94 & 0.08 && 76.22 & 0.07 && 0 \\
% Multi task &&& 36.35 & 0.26 && 0.6284 & 0.0034 && 21.02 & 0.06 && 76.36 & 0.05 && -1.89 \\
% MTI-Net~\cite{vandenhende2020mti} & \checkmark && 39.89 & 0.09 && \textbf{0.5824} & 0.0023 && 20.57 & 0.07 && \textbf{76.60} & 0.06 && 2.94 \\
% \hdashline\noalign{\vskip 0.5ex}
% ATRC (ours) &&& 38.90 & 0.43 && 0.6010 & 0.0046 && \textbf{20.48} & 0.02 && 76.34 & 0.12 && 1.56 \\
% & \checkmark && \textbf{40.80} & 0.34 && 0.5826 & 0.0020 && 20.51 & 0.05 && 76.50 & 0.11 && \textbf{3.57} \\
% \bottomrule
% \end{tabular}

% VERSION 2, WITHOUT STD
\begin{tabularx}{\textwidth}{@{}lcYYYYr@{}}\toprule
Model & FPM & SemSeg~$\uparrow$ & Depth~$\downarrow$ & Normal~$\downarrow$ & Bound~$\uparrow$ & $\Delta_m$~[\%]~$\uparrow$ \\ \midrule
Single task && 38.02 & 0.6104 & 20.94 & 76.22 & 0.00 \\
Multi-task && 36.35 & 0.6284 & 21.02 & 76.36 & -1.89 \\
MTI-Net~\cite{vandenhende2020mti} & \checkmark & 39.89 & 0.5824 & 20.57 & 76.60 & 2.94 \\
\hdashline\noalign{\vskip 0.5ex}
\multirow{ 2}{*}{ATRC (ours)} && 38.90 & 0.6010 & 20.48 & 76.34 & 1.56 \\
& \checkmark & 40.80 & 0.5826 & 20.51 & 76.50 & \textbf{3.57} \\
\bottomrule
\end{tabularx}%
}
\end{center}
\caption{NYUD-v2 performance comparison, using a HRNet18 backbone. FPM = Feature Propagation Module~\cite{vandenhende2020mti}.}
\label{tab:nyud18}
\end{table}

\begin{table}
\begin{center}
\resizebox{\linewidth}{!}{%

\ra{1.3}
\large
% \fontsize{6}{8}\selectfont

% \begin{tabular}{@{}lccrrcrrcrrcrrcr@{}}\toprule
% Model & FPM & \phantom{} & \multicolumn{2}{c}{SemSeg~$\uparrow$} & \phantom{a} & \multicolumn{2}{c}{Depth~$\downarrow$} & \phantom{a} & \multicolumn{2}{c}{Normal~$\downarrow$} & \phantom{a} & \multicolumn{2}{c}{Boundary~$\uparrow$} & \phantom{a} & $\Delta_m$~[\%]~$\uparrow$ \\
% \cmidrule{4-5} \cmidrule{7-8} \cmidrule{10-11} \cmidrule{13-14} &&& mean & std. && mean & std. && mean & std. && mean & std.\\ \midrule
% Single task &&& 45.87 & 0.12 && 0.5397 & 0.0033 && \textbf{20.09} & 0.07 && 77.34 & 0.16 && 0 \\
% Multi task &&& 41.96 & 0.62 && 0.5543 & 0.0056 && 20.36 & 0.13 && 77.62 & 0.12 && -3.05 \\
% MTI-Net~\cite{vandenhende2020mti} & \checkmark && 45.97 & 0.31 && 0.5365 & 0.0030 && 20.27 & 0.03 && 77.86 & 0.10 && 0.15 \\ \hdashline\noalign{\vskip 0.5ex}
% ATRC (ours) &&& 46.27 & 0.12 && 0.5495 & 0.0029 && 20.20 & 0.04 && 77.60 & 0.09 && -0.28 \\
% & \checkmark && \textbf{46.33} & 0.27 && \textbf{0.5363} & 0.0023 && 20.18 & 0.08 && \textbf{77.94} & 0.05 && \textbf{0.49} \\
% \bottomrule
% \end{tabular}

\begin{tabularx}{\textwidth}{@{}lcYYYYr@{}}\toprule
Model & FPM & SemSeg~$\uparrow$ & Depth~$\downarrow$ & Normal~$\downarrow$ & Bound~$\uparrow$ & $\Delta_m$~[\%]~$\uparrow$ \\ \midrule
Single task && 45.87 & 0.5397 & 20.09 & 77.34 & 0.00 \\
Multi-task && 41.96 & 0.5543 & 20.36 & 77.62 & -3.05 \\
MTI-Net~\cite{vandenhende2020mti} & \checkmark & 45.97 & 0.5365 & 20.27 & 77.86 & 0.15 \\ \hdashline\noalign{\vskip 0.5ex}
\multirow{ 2}{*}{ATRC (ours)} && 46.27 & 0.5495 & 20.20 & 77.60 & -0.28 \\
& \checkmark & 46.33 & 0.5363 & 20.18 & 77.94 & \textbf{0.49} \\
\bottomrule
\end{tabularx}%
}
\end{center}
\caption{NYUD-v2 performance comparison, using a HRNet48 backbone. FPM = Feature Propagation Module~\cite{vandenhende2020mti}.}
\label{tab:nyud48}
\end{table}

\begin{table}
\begin{center}
\resizebox{\linewidth}{!}{%

\ra{1.3}
\large

% \begin{tabular}{@{}lccrrcrrcrrcrrcrrcr@{}}\toprule
% Model & FPM & \phantom{} & \multicolumn{2}{c}{SemSeg~$\uparrow$} & \phantom{a} & \multicolumn{2}{c}{PartSeg~$\uparrow$} & \phantom{a} & \multicolumn{2}{c}{Sal~$\uparrow$} & \phantom{a} & \multicolumn{2}{c}{Normal~$\downarrow$} & \phantom{a} & \multicolumn{2}{c}{Boundary~$\uparrow$} & \phantom{a} & $\Delta_m$~[\%]~$\uparrow$ \\
% \cmidrule{4-5} \cmidrule{7-8} \cmidrule{10-11} \cmidrule{13-14} \cmidrule{16-17} &&& mean & std. && mean & std. && mean & std. && mean & std. && mean & std. \\ \midrule
% Single task &&& 62.23 & 0.24 && \textbf{61.66} & 0.17 && \textbf{85.08} & 0.07 && \textbf{13.69} & 0.05 && \textbf{73.06} & 0.05 && \textbf{0} \\
% Multi task &&& 51.48 & 0.23 && 57.23 & 0.12 && 83.43 & 0.10 && 14.10 & 0.03 && 69.76 & 0.08 && -6.77 \\
% MTI-Net~\cite{vandenhende2020mti} & \checkmark && 61.70 & 0.16 && 60.18 & 0.06 && 84.78 & 0.09 && 14.23 & 0.03 && 70.80 & 0.06 && -2.12 \\ \hdashline\noalign{\vskip 0.5ex}
% ATRC (ours) &&& 57.89 & 0.28 && 57.33 & 0.17 && 83.77 & 0.15 && 13.99 & 0.05 && 69.74 & 0.10 && -4.45 \\
% & \checkmark && \textbf{62.69} & 0.32 && 59.42 & 0.30 && 84.70 & 0.12 && 14.20 & 0.04 && 70.96 & 0.08 && -1.98 \\
% \bottomrule
% \end{tabular}

\begin{tabularx}{\textwidth}{@{}lcYYYYYr@{}}\toprule
Model & FPM & SemSeg~$\uparrow$ & PartSeg~$\uparrow$ & Sal~$\uparrow$ & Normal~$\downarrow$ & Bound~$\uparrow$ & $\Delta_m$~[\%]~$\uparrow$ \\ \midrule
Single task && 62.23 & 61.66 & 85.08 & 13.69 & 73.06 & \textbf{0.00} \\
Multi-task && 51.48 & 57.23 & 83.43 & 14.10 & 69.76 & -6.77 \\
MTI-Net~\cite{vandenhende2020mti} & \checkmark & 61.70 & 60.18 & 84.78 & 14.23 & 70.80 & -2.12 \\ \hdashline\noalign{\vskip 0.5ex}
\multirow{ 2}{*}{ATRC (ours)} && 57.89 & 57.33 & 83.77 & 13.99 & 69.74 & -4.45 \\
& \checkmark & 62.69 & 59.42 & 84.70 & 14.20 & 70.96 & -1.98 \\
\bottomrule
\end{tabularx}%
}
\end{center}
\caption{PASCAL-Context performance comparison, using a HRNet18 backbone. FPM = Feature Propagation Module~\cite{vandenhende2020mti}.}
\label{tab:pascal18}
\end{table}

\subsection{Source Task Importance}
\label{subsec:importance}

The simple design of the proposed ATRC module allows us to investigate the importance of each source-target task connection ($\widehat{=}$ CP block) for the final predictions of fitted models. 
To this end, we adapt permutation feature importance~\cite{breiman2001random} to our setting. 
We can determine the importance of a CP block by recording the drop in multi-task performance $\Delta_m$ when the output of that block is randomly shuffled over the dataset. 
To get a more reliable estimate, this procedure is repeated multiple times with different permutations.
% Using the test dataset, we store the output features of each CP block and repeat the following two steps multiple times for each CP block: 
% (1) Permute output features of the inspected CP block over the test samples. 
% (2) Evaluate this model on the test dataset and record the drop in multi-task performance $\Delta_m$. 
Neglecting feature multicollinearity, the average drop in $\Delta_m$ provides an estimate of how strongly the fitted model depends on the inspected source task for the corresponding target task prediction.
We use held-out data in this experiment to assess the importance for generalization power.

\begin{figure}[t]
    \begin{minipage}[c]{0.48\linewidth}
        \includegraphics[width=\linewidth]{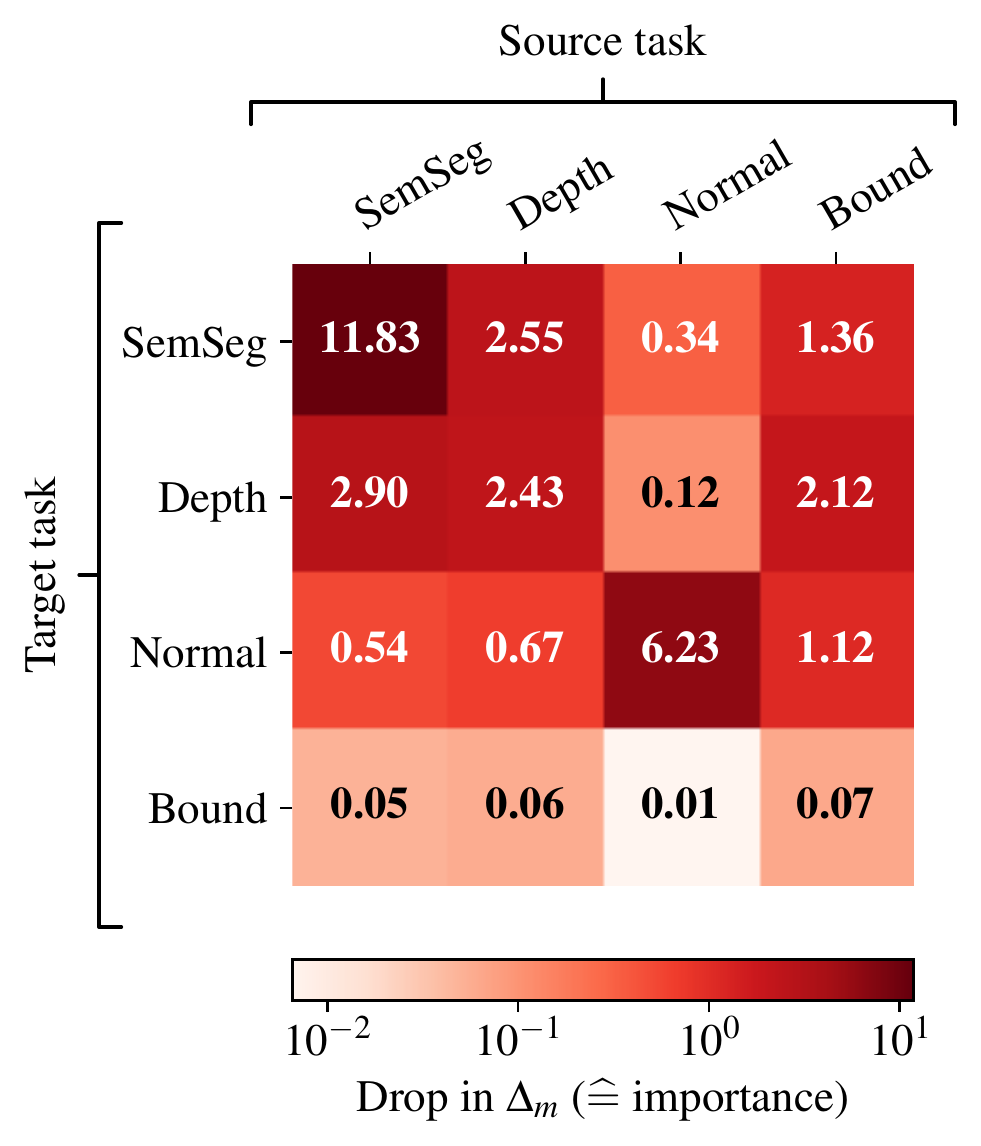}
    \end{minipage}\hfill
    \begin{minipage}[c]{0.49\linewidth}
        \caption{Source task importance; measured by permutation testing of fitted ATRC models on NYUD-v2. The contribution of a source task in the distillation is gauged by the drop in multi-task performance $\Delta_m$ as the output of the corresponding source-target task distillation is randomly permuted. The values shown in the matrix are mean percentage drops in $\Delta_m$.}
        \label{fig:comp}
    \end{minipage}
\end{figure}

Fig.~\ref{fig:comp} visualizes the results for NYUD-v2. The inspection reveals that self-attention remains the most important distillation connection for three out of four tasks. However, depth estimation seems to rely more strongly on semantic segmentation source features, corroborating empirical evidence in the literature that depth estimation can be improved significantly using semantic predictions~\cite{xu2018pad}. Overall, boundary detection profits little from multi-modal distillation according to this analysis, which is consistent with the lack of noteworthy performance gain for this task in Table~\ref{tab:control}.
We hypothesize that this could be due to the large discrepancy between the loss (we follow others~\cite{maninis2019attentive,vandenhende2020mti,kanakis2020reparameterizing} and use balanced cross entropy) and metric for this task. A more tailored loss function such as~\cite{kokkinos2015pushing} might help in this case. 
% however this is beyond the scope of our work.

Source task importance scores are linearly correlated with the search algorithm reliability---albeit weakly (Pearson correlation coefficient of 0.43). 
Notably, we observe 100\% reliability for the three most important source-target task connections of Fig.~\ref{fig:comp}.
This suggests that the search algorithm is more consistent for important decisions.
We quantify search algorithm reliability using percentage agreement in candidate selection between all search run pairs (does not account for chance agreement, see Appendix~\ref{sec:reliability}). 

% \begin{figure}[t]
% \centering
% \includegraphics[width=0.8\linewidth]{figs/importance.pdf}
%    \caption{Task importance as measured by permutation testing of trained ATRC models on NYUD-v2. The contribution of a source task in the distillation is measured by permuting distillation module outputs of this source task over the test dataset while recording the drop in multi-task performance $\Delta_m$. The values shown in the matrix are mean percentage drops in $\Delta_m$.}
% \label{fig:comp}
% \end{figure}

\subsection{Complementary Methods}

To demonstrate its flexibility, we combine our ATRC module with (1) the contextual Atrous Spatial Pyramid Pooling (ASPP) module of~\cite{chen2018encoder} and (2) automatic backbone branching via Branched Multi Task Architecture Search (BMTAS)~\cite{bruggemann2020automated}.
For these experiments, we use a dilated ResNet-50 backbone (output stride 16) with a skip connection at stride 4, and fully convolutional task-specific heads.

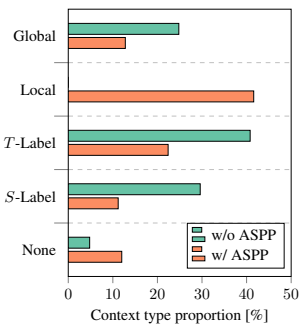
\begin{figure}[t]
    \begin{minipage}[c]{0.5\linewidth}
        \resizebox{\linewidth}{!}{%
            % This file was created by tikzplotlib v0.9.8.
\begin{tikzpicture}

\definecolor{color0}{rgb}{0.4,0.76078431372549,0.647058823529412}
\definecolor{color1}{rgb}{0.988235294117647,0.552941176470588,0.384313725490196}

\begin{axis}[
    scale only axis=true,
    height=6cm,
    width=5cm,
    xbar,
    bar width=7pt,
    tick align=outside,
    tick pos=left,
    ytick = {1,2,3,4,5},
    yticklabels={None,$S$-Label,$T$-Label,Local,Global},
    ymin=0.5, ymax=5.5,
    y grid style=dashed,
    yminorgrids,
    xlabel={Context type proportion [\%]},
    xmin=0, xmax=50,
    major y tick style = {opacity=0},
    minor y tick num = 1,
    minor tick length=2ex,
    reverse legend,
    legend cell align={left},
	legend style={column sep=2pt},
	legend pos=south east
]

\addplot[draw=black,fill=color1]
	coordinates {
	(12.0,1)
	(11.2,2)
	(22.4,3)
	(41.6,4)
	(12.8,5)
	};
\addplot[draw=black,fill=color0] 
	coordinates {
	(4.8,1)
	(29.6,2)
	(40.8,3)
	(0.0,4)
	(24.8,5)
	};
\legend{w/ ASPP,w/o ASPP}

\end{axis}

\end{tikzpicture}
        }%
    \end{minipage}\hfill
    \begin{minipage}[c]{0.49\linewidth}
       \caption{Proportions of selected context types over five search runs, for architectures without and with an ASPP module~\cite{chen2018encoder} inserted before the ATRC module. The change in proportion of the local context indicates that ATRC adapts to better complement the new backbone. This experiment was conducted on the PASCAL-Context dataset using a ResNet-50 based architecture.}
       \label{fig:pie}
    \end{minipage}
\end{figure}

ASPP is a popular multi-scale context aggregation module leveraging dilated convolutions.
We insert a separate ASPP module before each task-specific head. 
Table~\ref{tab:aspp} shows that ATRC also improves the performance of the ASPP-augmented network, indicating that the two context aggregation stages are complementary to some extent.
Interestingly, the proportions of selected relational context types in the ATRC search change drastically with ASPP, as illustrated in Fig.~\ref{fig:pie}: 
The proportion of local context rises from 0\% (w/o ASPP) to 41.6\% (w/ ASPP), demonstrating that ATRC adapts the context types given the nature of different backbones (\eg, the enhanced receptive field of ASPP is better complemented with local information).
% suggesting that the multi-scale aggregation effect of the ASPP can be partially recovered using relational context with a global receptive field.

Branched networks are a hard parameter sharing MTL strategy and, as such, complementary to multi-modal distillation (see Sec.~\ref{sec:related}). We show this by applying our method in combination with a branched backbone configuration, determined through the NAS-based BMTAS. The results in Table~\ref{tab:aspp} demonstrate that ATRC improves performance also for branched multi-task networks. 

\begin{table}
\begin{center}
\resizebox{\linewidth}{!}{%

\ra{1.3}
\large

\begin{tabularx}{\textwidth}{@{}lcYYYYYr@{}}\toprule
Model & ATRC & SemSeg~$\uparrow$ & PartSeg~$\uparrow$ & Sal~$\uparrow$ & Normal~$\downarrow$ & Bound~$\uparrow$ & $\Delta_m$~[\%]~$\uparrow$ \\ \midrule
Single task && 56.65 & 62.67 & 80.62 & 14.66 & 74.00 & 0.00 \\
% & \checkmark & 69.61 & 62.99 & 84.57 & 13.92 & 74.08 & 6.69 \\
\hdashline\noalign{\vskip 0.5ex}
\multirow{ 2}{*}{Multi-task} && 50.78 & 59.37 & 78.99 & 15.16 & 71.18 & -4.97 \\
& \checkmark & 62.99 & 59.79 & 82.25 & 14.67 & 71.20 & \textbf{0.95} \\
\hdashline\noalign{\vskip 0.5ex}
\multirow{ 2}{*}{ASPP~\cite{chen2018encoder}} && 62.70 & 59.98 & 83.81 & 14.34 & 71.28 & 1.77 \\
& \checkmark & 63.60 & 60.23 & 83.91 & 14.30 & 70.86 & \textbf{2.13} \\
\hdashline\noalign{\vskip 0.5ex}
\multirow{ 2}{*}{BMTAS~\cite{bruggemann2020automated}} && 56.37 & 62.54 & 79.91 & 14.60 & 72.83 & -0.55 \\
& \checkmark & 67.67 & 62.93 & 82.29 & 14.24 & 72.42 & \textbf{4.53} \\
\bottomrule
\end{tabularx}%
}
\end{center}
\caption{PASCAL-Context performance of ASPP~\cite{chen2018encoder} and BMTAS~\cite{bruggemann2020automated} when supplemented with our ATRC. For ASPP, we insert an ASPP module at the beginning of each task-specific head. For BMTAS, we use their method to find a branched backbone (instead of fully shared). ATRC is complementary to both approaches. Experiments are based on a dilated ResNet-50 backbone.}
\label{tab:aspp}
\end{table}

% \begin{table}
% \begin{center}
% \resizebox{\linewidth}{!}{%
% \input{tables/pascal_comp2.tex}%
% }
% \end{center}
% \caption{PASCAL-Context performance of ATRC applied to a branched backbone. We determine the branching configuration automatically using BMTAS~\cite{bruggemann2020automated}, based on a ResNet-50 backbone.}
% \label{tab:bmtas}
% \end{table}

% \begin{figure}[t]
% \centering
% \resizebox{\linewidth}{!}{%
% \input{figs/pie.tex}
% }%
%    \caption{Proportions of selected context types over five search runs, for architectures without (left) and with (right) an ASPP module~\cite{chen2018encoder} inserted before the ATRC module. The change in proportion of the local context indicates that ATRC adapts to better complement the new backbone.}
% \label{fig:pie}
% \end{figure}

\section{Conclusion}
\label{sec:conclusion}

We presented ATRC, a novel multi-modal distillation module which exploits inter- and intra-task relationships to refine pixel-wise predictions. 
The proposed approach leverages scaled dot-product attention to enrich the features of a target task through contextual source task features, while explicitly factoring in tasks' relations. 
We formulate four relational context types for multi-modal distillation (\emph{global}, \emph{local}, \emph{T-label}, and \emph{S-label} context) and detail an algorithm which customizes the context type for every given source-target task pair.  %, while jointly optimizing the cross-task information exchange. 
Experimental analyses on NYUD-v2 and PASCAL-Context benchmarks indicate that our ATRC module outperforms comparable multi-modal distillation modules established in the literature. 
Overall, the presented framework shows great promise for multi-task dense prediction and opens the door for future research in customized task-relational context descriptions.

{\small
\bibliographystyle{ieee_fullname}
\bibliography{egbib}

\begin{thebibliography}{10}\itemsep=-1pt

\bibitem{achanta2009frequency}
Radhakrishna Achanta, Sheila Hemami, Francisco Estrada, and Sabine Susstrunk.
\newblock Frequency-tuned salient region detection.
\newblock In {\em CVPR}, 2009.

\bibitem{bahdanau2014neural}
Dzmitry Bahdanau, Kyunghyun Cho, and Yoshua Bengio.
\newblock Neural machine translation by jointly learning to align and
  translate.
\newblock In {\em ICLR}, 2015.

\bibitem{bello2019attention}
Irwan Bello, Barret Zoph, Ashish Vaswani, Jonathon Shlens, and Quoc~V Le.
\newblock Attention augmented convolutional networks.
\newblock In {\em ICCV}, 2019.

\bibitem{breiman2001random}
Leo Breiman.
\newblock Random forests.
\newblock {\em Machine learning}, 45(1):5--32, 2001.

\bibitem{bruggemann2020automated}
David Bruggemann, Menelaos Kanakis, Stamatios Georgoulis, and Luc Van~Gool.
\newblock Automated search for resource-efficient branched multi-task networks.
\newblock In {\em BMVC}, 2020.

\bibitem{Caruana93multitasklearning:}
Richard Caruana.
\newblock Multitask learning: A knowledge-based source of inductive bias.
\newblock In {\em ICML}, 1993.

\bibitem{chen2017deeplab}
Liang-Chieh Chen, George Papandreou, Iasonas Kokkinos, Kevin Murphy, and Alan~L
  Yuille.
\newblock Deeplab: Semantic image segmentation with deep convolutional nets,
  atrous convolution, and fully connected crfs.
\newblock {\em TPAMI}, 40(4):834--848, 2017.

\bibitem{chen2018encoder}
Liang-Chieh Chen, Yukun Zhu, George Papandreou, Florian Schroff, and Hartwig
  Adam.
\newblock Encoder-decoder with atrous separable convolution for semantic image
  segmentation.
\newblock In {\em ECCV}, 2018.

\bibitem{chen2014detect}
Xianjie Chen, Roozbeh Mottaghi, Xiaobai Liu, Sanja Fidler, Raquel Urtasun, and
  Alan Yuille.
\newblock Detect what you can: Detecting and representing objects using
  holistic models and body parts.
\newblock In {\em CVPR}, 2014.

\bibitem{chen20182}
Yunpeng Chen, Yannis Kalantidis, Jianshu Li, Shuicheng Yan, and Jiashi Feng.
\newblock $a^2$-nets: Double attention networks.
\newblock In {\em NeurIPS}, 2018.

\bibitem{chen2018gradnorm}
Zhao Chen, Vijay Badrinarayanan, Chen-Yu Lee, and Andrew Rabinovich.
\newblock Gradnorm: Gradient normalization for adaptive loss balancing in deep
  multitask networks.
\newblock In {\em ICML}, 2018.

\bibitem{cohen1960coefficient}
Jacob Cohen.
\newblock A coefficient of agreement for nominal scales.
\newblock {\em Educational and psychological measurement}, 20(1):37--46, 1960.

\bibitem{dosovitskiy2021an}
Alexey Dosovitskiy, Lucas Beyer, Alexander Kolesnikov, Dirk Weissenborn,
  Xiaohua Zhai, Thomas Unterthiner, Mostafa Dehghani, Matthias Minderer, Georg
  Heigold, Sylvain Gelly, Jakob Uszkoreit, and Neil Houlsby.
\newblock An image is worth 16x16 words: Transformers for image recognition at
  scale.
\newblock In {\em ICLR}, 2021.

\bibitem{everingham2010pascal}
Mark Everingham, Luc Van~Gool, Christopher~KI Williams, John Winn, and Andrew
  Zisserman.
\newblock The pascal visual object classes (voc) challenge.
\newblock {\em IJCV}, 88(2):303--338, 2010.

\bibitem{fu2019dual}
Jun Fu, Jing Liu, Haijie Tian, Yong Li, Yongjun Bao, Zhiwei Fang, and Hanqing
  Lu.
\newblock Dual attention network for scene segmentation.
\newblock In {\em CVPR}, 2019.

\bibitem{gao2020mtl}
Yuan Gao, Haoping Bai, Zequn Jie, Jiayi Ma, Kui Jia, and Wei Liu.
\newblock Mtl-nas: Task-agnostic neural architecture search towards
  general-purpose multi-task learning.
\newblock In {\em CVPR}, 2020.

\bibitem{gao2019nddr}
Yuan Gao, Jiayi Ma, Mingbo Zhao, Wei Liu, and Alan~L Yuille.
\newblock Nddr-cnn: Layerwise feature fusing in multi-task cnns by neural
  discriminative dimensionality reduction.
\newblock In {\em CVPR}, 2019.

\bibitem{gupta2014learning}
Saurabh Gupta, Ross Girshick, Pablo Arbel{\'a}ez, and Jitendra Malik.
\newblock Learning rich features from rgb-d images for object detection and
  segmentation.
\newblock In {\em ECCV}, 2014.

\bibitem{he2016deep}
Kaiming He, Xiangyu Zhang, Shaoqing Ren, and Jian Sun.
\newblock Deep residual learning for image recognition.
\newblock In {\em CVPR}, 2016.

\bibitem{hu2020dsnas}
Shoukang Hu, Sirui Xie, Hehui Zheng, Chunxiao Liu, Jianping Shi, Xunying Liu,
  and Dahua Lin.
\newblock Dsnas: Direct neural architecture search without parameter
  retraining.
\newblock In {\em CVPR}, 2020.

\bibitem{jacobs1991adaptive}
Robert~A Jacobs, Michael~I Jordan, Steven~J Nowlan, and Geoffrey~E Hinton.
\newblock Adaptive mixtures of local experts.
\newblock {\em Neural computation}, 3(1):79--87, 1991.

\bibitem{jang2016categorical}
Eric Jang, Shixiang Gu, and Ben Poole.
\newblock Categorical reparameterization with gumbel-softmax.
\newblock In {\em ICLR}, 2017.

\bibitem{kanakis2020reparameterizing}
Menelaos Kanakis, David Bruggemann, Suman Saha, Stamatios Georgoulis, Anton
  Obukhov, and Luc Van~Gool.
\newblock Reparameterizing convolutions for incremental multi-task learning
  without task interference.
\newblock In {\em ECCV}, 2020.

\bibitem{katharopoulos2020transformers}
Angelos Katharopoulos, Apoorv Vyas, Nikolaos Pappas, and Fran{\c{c}}ois
  Fleuret.
\newblock Transformers are rnns: Fast autoregressive transformers with linear
  attention.
\newblock In {\em ICML}, 2020.

\bibitem{kendall2018multi}
Alex Kendall, Yarin Gal, and Roberto Cipolla.
\newblock Multi-task learning using uncertainty to weigh losses for scene
  geometry and semantics.
\newblock In {\em CVPR}, 2018.

\bibitem{kingma2014adam}
Diederik~P Kingma and Jimmy Ba.
\newblock Adam: A method for stochastic optimization.
\newblock In {\em ICLR}, 2015.

\bibitem{kokkinos2015pushing}
Iasonas Kokkinos.
\newblock Pushing the boundaries of boundary detection using deep learning.
\newblock In {\em ICLR}, 2016.

\bibitem{kokkinos2017ubernet}
Iasonas Kokkinos.
\newblock Ubernet: Training a universal convolutional neural network for low-,
  mid-, and high-level vision using diverse datasets and limited memory.
\newblock In {\em CVPR}, 2017.

\bibitem{krizhevsky2012imagenet}
Alex Krizhevsky, Ilya Sutskever, and Geoffrey~E Hinton.
\newblock Imagenet classification with deep convolutional neural networks.
\newblock In {\em NeurIPS}, 2012.

\bibitem{lecun1998gradient}
Yann LeCun, Léon Bottou, Yoshua Bengio, and Patrick Haffner.
\newblock Gradient-based learning applied to document recognition.
\newblock {\em Proceedings of the IEEE}, 86(11):2278--2324, 1998.

\bibitem{li2018monocular}
Bo Li, Yuchao Dai, and Mingyi He.
\newblock Monocular depth estimation with hierarchical fusion of dilated cnns
  and soft-weighted-sum inference.
\newblock {\em Pattern Recognition}, 83:328--339, 2018.

\bibitem{light1971measures}
Richard~J Light.
\newblock Measures of response agreement for qualitative data: some
  generalizations and alternatives.
\newblock {\em Psychological bulletin}, 76(5):365, 1971.

\bibitem{liu2018darts}
Hanxiao Liu, Karen Simonyan, and Yiming Yang.
\newblock Darts: Differentiable architecture search.
\newblock In {\em ICLR}, 2019.

\bibitem{liu2019end}
Shikun Liu, Edward Johns, and Andrew~J Davison.
\newblock End-to-end multi-task learning with attention.
\newblock In {\em CVPR}, 2019.

\bibitem{liu1999ensemble}
Yong Liu and Xin Yao.
\newblock Ensemble learning via negative correlation.
\newblock {\em Neural networks}, 12(10):1399--1404, 1999.

\bibitem{lowe1999sift}
David~G Lowe.
\newblock Object recognition from local scale-invariant features.
\newblock In {\em ICCV}, 1999.

\bibitem{lu2017fully}
Yongxi Lu, Abhishek Kumar, Shuangfei Zhai, Yu Cheng, Tara Javidi, and Rogerio
  Feris.
\newblock Fully-adaptive feature sharing in multi-task networks with
  applications in person attribute classification.
\newblock In {\em CVPR}, 2017.

\bibitem{maddison2016concrete}
Chris~J Maddison, Andriy Mnih, and Yee~Whye Teh.
\newblock The concrete distribution: A continuous relaxation of discrete random
  variables.
\newblock In {\em ICLR}, 2017.

\bibitem{maninis2019attentive}
Kevis-Kokitsi Maninis, Ilija Radosavovic, and Iasonas Kokkinos.
\newblock Attentive single-tasking of multiple tasks.
\newblock In {\em CVPR}, 2019.

\bibitem{martin2004learning}
David~R Martin, Charless~C Fowlkes, and Jitendra Malik.
\newblock Learning to detect natural image boundaries using local brightness,
  color, and texture cues.
\newblock {\em TPAMI}, 26(5):530--549, 2004.

\bibitem{misra2016cross}
Ishan Misra, Abhinav Shrivastava, Abhinav Gupta, and Martial Hebert.
\newblock Cross-stitch networks for multi-task learning.
\newblock In {\em CVPR}, 2016.

\bibitem{ramachandran2019stand}
Prajit Ramachandran, Niki Parmar, Ashish Vaswani, Irwan Bello, Anselm Levskaya,
  and Jonathon Shlens.
\newblock Stand-alone self-attention in vision models.
\newblock In {\em NeurIPS}, 2019.

\bibitem{real2019regularized}
Esteban Real, Alok Aggarwal, Yanping Huang, and Quoc~V Le.
\newblock Regularized evolution for image classifier architecture search.
\newblock In {\em AAAI}, 2019.

\bibitem{real2017large}
Esteban Real, Sherry Moore, Andrew Selle, Saurabh Saxena, Yutaka~Leon Suematsu,
  Jie Tan, Quoc~V Le, and Alexey Kurakin.
\newblock Large-scale evolution of image classifiers.
\newblock In {\em ICML}, 2017.

\bibitem{sener2018multi}
Ozan Sener and Vladlen Koltun.
\newblock Multi-task learning as multi-objective optimization.
\newblock In {\em NeurIPS}, 2018.

\bibitem{silberman2012indoor}
Nathan Silberman, Derek Hoiem, Pushmeet Kohli, and Rob Fergus.
\newblock Indoor segmentation and support inference from rgbd images.
\newblock In {\em ECCV}, 2012.

\bibitem{simonyan2015very}
Karen Simonyan and Andrew Zisserman.
\newblock Very deep convolutional networks for large-scale image recognition.
\newblock In {\em ICLR}, 2015.

\bibitem{su2019pixel}
Hang Su, Varun Jampani, Deqing Sun, Orazio Gallo, Erik Learned-Miller, and Jan
  Kautz.
\newblock Pixel-adaptive convolutional neural networks.
\newblock In {\em CVPR}, 2019.

\bibitem{vandenhende2019branched}
Simon Vandenhende, Stamatios Georgoulis, Bert De~Brabandere, and Luc Van~Gool.
\newblock Branched multi-task networks: deciding what layers to share.
\newblock In {\em BMVC}, 2020.

\bibitem{vandenhende2020mti}
Simon Vandenhende, Stamatios Georgoulis, and Luc Van~Gool.
\newblock Mti-net: Multi-scale task interaction networks for multi-task
  learning.
\newblock In {\em ECCV}, 2020.

\bibitem{vaswani2017attention}
Ashish Vaswani, Noam Shazeer, Niki Parmar, Jakob Uszkoreit, Llion Jones,
  Aidan~N Gomez, {\L}ukasz Kaiser, and Illia Polosukhin.
\newblock Attention is all you need.
\newblock In {\em NeurIPS}, 2017.

\bibitem{wang2020deep}
Jingdong Wang, Ke Sun, Tianheng Cheng, Borui Jiang, Chaorui Deng, Yang Zhao,
  Dong Liu, Yadong Mu, Mingkui Tan, Xinggang Wang, et~al.
\newblock Deep high-resolution representation learning for visual recognition.
\newblock {\em TPAMI}, 2020.

\bibitem{wang2018non}
Xiaolong Wang, Ross Girshick, Abhinav Gupta, and Kaiming He.
\newblock Non-local neural networks.
\newblock In {\em CVPR}, 2018.

\bibitem{xie2017aggregated}
Saining Xie, Ross Girshick, Piotr Doll{\'a}r, Zhuowen Tu, and Kaiming He.
\newblock Aggregated residual transformations for deep neural networks.
\newblock In {\em CVPR}, 2017.

\bibitem{xie2018snas}
Sirui Xie, Hehui Zheng, Chunxiao Liu, and Liang Lin.
\newblock Snas: Stochastic neural architecture search.
\newblock In {\em ICLR}, 2019.

\bibitem{xu2018pad}
Dan Xu, Wanli Ouyang, Xiaogang Wang, and Nicu Sebe.
\newblock Pad-net: Multi-tasks guided prediction-and-distillation network for
  simultaneous depth estimation and scene parsing.
\newblock In {\em CVPR}, 2018.

\bibitem{xu2015show}
Kelvin Xu, Jimmy Ba, Ryan Kiros, Kyunghyun Cho, Aaron Courville, Ruslan
  Salakhudinov, Rich Zemel, and Yoshua Bengio.
\newblock Show, attend and tell: Neural image caption generation with visual
  attention.
\newblock In {\em ICML}, 2015.

\bibitem{yin2020disentangled}
Minghao Yin, Zhuliang Yao, Yue Cao, Xiu Li, Zheng Zhang, Stephen Lin, and Han
  Hu.
\newblock Disentangled non-local neural networks.
\newblock In {\em ECCV}, 2020.

\bibitem{yuan2019object}
Yuhui Yuan, Xilin Chen, and Jingdong Wang.
\newblock Object-contextual representations for semantic segmentation.
\newblock In {\em ECCV}, 2020.

\bibitem{zeisl2014discriminatively}
Bernhard Zeisl, Marc Pollefeys, et~al.
\newblock Discriminatively trained dense surface normal estimation.
\newblock In {\em ECCV}, 2014.

\bibitem{zhang2019acfnet}
Fan Zhang, Yanqin Chen, Zhihang Li, Zhibin Hong, Jingtuo Liu, Feifei Ma, Junyu
  Han, and Errui Ding.
\newblock Acfnet: Attentional class feature network for semantic segmentation.
\newblock In {\em ICCV}, 2019.

\bibitem{zhang2019co}
Hang Zhang, Han Zhang, Chenguang Wang, and Junyuan Xie.
\newblock Co-occurrent features in semantic segmentation.
\newblock In {\em CVPR}, 2019.

\bibitem{zhang2019pattern}
Zhenyu Zhang, Zhen Cui, Chunyan Xu, Yan Yan, Nicu Sebe, and Jian Yang.
\newblock Pattern-affinitive propagation across depth, surface normal and
  semantic segmentation.
\newblock In {\em CVPR}, 2019.

\bibitem{zhou2020pattern}
Ling Zhou, Zhen Cui, Chunyan Xu, Zhenyu Zhang, Chaoqun Wang, Tong Zhang, and
  Jian Yang.
\newblock Pattern-structure diffusion for multi-task learning.
\newblock In {\em CVPR}, 2020.

\bibitem{zoph2016neural}
Barret Zoph and Quoc~V. Le.
\newblock Neural architecture search with reinforcement learning.
\newblock In {\em ICLR}, 2017.

\bibitem{zoph2018learning}
Barret Zoph, Vijay Vasudevan, Jonathon Shlens, and Quoc~V Le.
\newblock Learning transferable architectures for scalable image recognition.
\newblock In {\em CVPR}, 2018.

\end{thebibliography}
}

%\newpage

\begin{appendices}

\beginappendixa
\section{Training Details}
\label{sec:train_details}

In this section, we describe the training setup. 
For consistency, all experiments in the paper were repeated five times using the pipeline detailed below.
% {\tt PyTorch} training and evaluation code for this project is available at \mbox{\url{https://github.com/brdav/atrc}}.

\textbf{Data augmentation.} We augment input images during training by random scaling with values between 0.5 and 2.0, random cropping to input size ($425 \times 560$ for NYUD-v2---we use the cropped version of~\cite{gupta2014learning}---and padded to $512 \times 512$ for PASCAL-Context), random horizontal flipping and random color jitter. Image intensities are standardized. Depth labels are corrected for scaling and surface normal labels are corrected for horizontal flipping.

\textbf{Task losses.}
The total loss of the multi-task network with parameters $\theta$ is a weighted sum of losses (for tasks $n \in \{1,...,N\}$):
\begin{equation}
    \mathcal{L}_{total}(\theta) = \sum_{n=1}^N \omega_n \mathcal{L}_n(\theta)
\label{eq:weighted_sum}
\end{equation}
For semantic segmentation and human parts segmentation we use a cross-entropy loss (loss weights $\omega_n = 1$ and $\omega_n = 2$ respectively), for saliency estimation a balanced cross-entropy loss ($\omega_n = 5$), for depth estimation a $\mathcal{L}_1$ loss ($\omega_n = 1$), for surface normal estimation a $\mathcal{L}_1$ loss with unit vector normalization ($\omega_n = 10$) and for boundary detection a weighted cross-entropy loss ($\omega_n = 50$). For boundary detection, the positive pixels are weighted with 0.8 and the negative pixels with 0.2 on NYUD-v2, while on PASCAL-Context the weights are 0.95 and 0.05. $\omega_n$ for each task was determined through a logarithmic grid search over candidate values with single-task networks.

The auxiliary predictions $A_n$ are trained with a cross-entropy loss using the same loss weights as above. However, the auxiliary head backpropagation is stopped from updating parameters of the main network.

\textbf{Optimization hyperparameters.} 
All backbones are initialized with ImageNet pretrained weights. We use Stochastic Gradient Descent (SGD) with momentum of 0.9 and weight decay of 0.0005 to optimize the model parameters. The initial learning rate is determined through a logarithmic grid search (..., 0.002, 0.005, 0.01, 0.02, ...), with the option of having a 10 times higher learning rate for the heads \vs the backbone. The initial value is decayed during training according to a `poly' learning rate schedule~\cite{chen2017deeplab}. For all experiments, we use a minibatch size of 8 and train for 40000 iterations.

\textbf{Context type search.}
The architecture distribution parameters $\alpha$ are initialized with zeros.
We use an Adam optimizer~\cite{kingma2014adam} to update them, with learning rate 0.0005 (no weight decay, no learning rate scheduler). The update occurs in the same round of backpropagation as the regular model parameters (single-level optimization). 
Over the course of training, the Gumbel-Softmax temperature $\lambda$ is annealed linearly from 1.0 to 0.05 (following~\cite{xie2018snas}). 
Also, to ensure a fair candidate context type selection, we disable learnable affine parameters of the last batch normalization of every context type attention mechanism.

As discussed in Sec.~\ref{subsec:auto}, we use entropy ($H$) regularization to control the sampling variance during the architecture search.
Specifically, we calculate the mean entropy of the architecture parameter ($\alpha$)-distributions over all Context Pooling (CP) blocks, scale it with a weight $\omega_{H}$, and add it to the total loss.
\begin{equation}
    \mathcal{L}_{search}(\theta,\alpha) = \sum_{n=1}^N \omega_n \mathcal{L}_n(\theta,\alpha) + \frac{\omega_{H}}{N^2}\sum_{j=1}^{N^2} H(\alpha_j)
\label{eq:search_weighted_sum}
\end{equation}
$j$ indexes the CP blocks. The scaling factor $\omega_H$ follows a linear schedule during the search, from -0.02 to 0.06. We found that this provides an adequate balance between candidate exploration and exploitation.
For a given CP block $j$, architecture search is terminated prematurely if the difference between the two largest values of $\alpha_j$ exceeds 0.3.
One candidate is then sampled using $\argmax$ (\ie, $\alpha_j$ becomes a one-hot vector).

After concluding five runs of the architecture search, we determine the final configuration by choosing the context type receiving the most votes over the five runs in each CP block. Ultimately, this final configuration is retrained five times.

\beginappendixb
\section{Implementation Verification}

We verify the implementation of our pipeline by comparing HRNet single task performances with the numbers published in~\cite{vandenhende2020mti}.
Table~\ref{tab:stcomp} shows that the baselines trained with our pipeline outperform those of~\cite{vandenhende2020mti}.

For implementing the various distillation modules in Table~\ref{tab:control}, we used the code provided by the authors whenever possible, and otherwise followed the information provided in the papers closely. For MTI-Net~\cite{vandenhende2020mti}, we used the authors' model code within our pipeline. 

Finally, we attempted to reimplement the full PSD~\cite{zhou2020pattern} network based on a ResNet-50 backbone (as suggested in the original paper), but were unable to obtain competitive results.

\begin{table}
\begin{center}
\resizebox{\linewidth}{!}{%
\ra{1.3}
\Large
\begin{tabularx}{\textwidth}{@{}lYYYr@{}}\toprule
Model & SemSeg~$\uparrow$ & Depth~$\downarrow$ & Normal~$\downarrow$ & Bound~$\uparrow$ \\ \midrule
HRNet18, \cite{vandenhende2020mti} & 33.18 & 0.667 & - & - \\
HRNet18, ours & 38.02 & 0.610 & 20.94 & 76.22 \\
\hdashline\noalign{\vskip 0.5ex}
HRNet48, \cite{vandenhende2020mti} & 45.70 & 0.547 & - & - \\
HRNet48, ours & 45.87 & 0.540 & 20.09 & 77.34 \\
\bottomrule
\end{tabularx}%
}
\end{center}
\caption{NYUD-v2 single task performances of HRNetV2-W18-small (HRNet18) and HRNetV2-W48 (HRNet48) models~\cite{wang2020deep}. We compare the performances obtained using our implementation with the numbers published in \cite{vandenhende2020mti}.}
\label{tab:stcomp}
\end{table}

\beginappendixc
\section{Relational Context Schematics}

\begin{figure*}[t]
    \includegraphics[width=\linewidth]{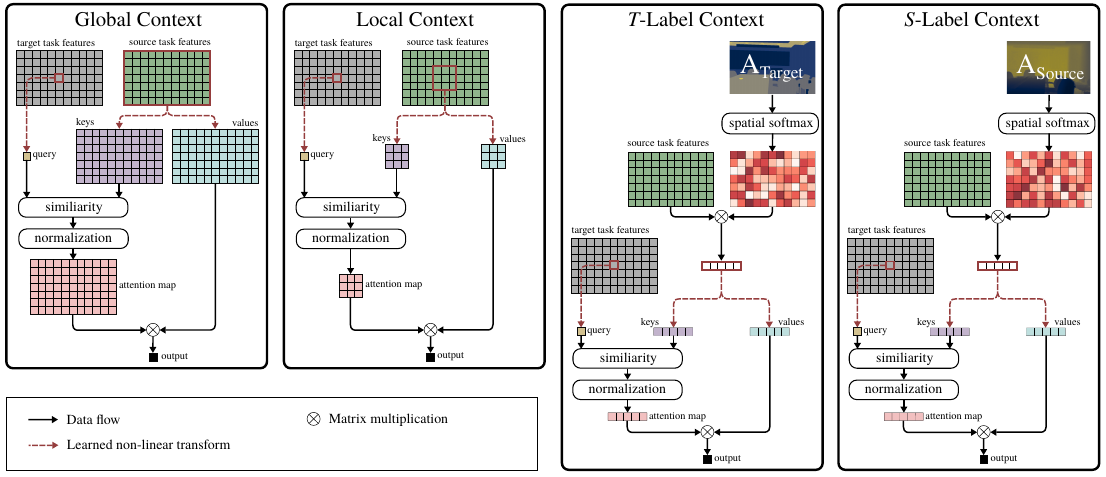}
    \caption{Schematics of the different relational context types. The grids represent individual pixels (channels not shown), the attention mechanism is shown for one target pixel (framed in red box) respectively. Normalization is applied over all pixels of the attention map. $A_*$ are the auxiliary predictions, as depicted in Fig.~\ref{fig:svd_network}.}
    \label{fig:overview}
\end{figure*}

Fig.~\ref{fig:overview} depicts the different relational context types used in this work schematically. We use a $\oneconv$-$\batchnorm$-$\relu$ layer as the learned non-linear transform. For all contexts except global, the similarity function is $\similarity(q_i, k_j) = \exp(\frac{q_i k_j^\intercal}{d_k})$ (which corresponds to $\softmax$). For the global context, it is simply $\similarity(q_i, k_j) = q_i k_j^\intercal$.

\beginappendixd
\section{Label Context: Regression Tasks}
\label{sec:regression_tasks}

In this section, we discuss how the label space of regression tasks can be partitioned into distinct regions for label context formation, as mentioned in Sec.~\ref{subsec:similarity_context}. Regression tasks can be easily reformulated as classification tasks by binning the continuous ground truth values. However, the discretization scheme has to be tailored towards each task separately to obtain satisfactory performance.

For depth prediction, our approach is inspired by~\cite{li2018monocular}. Specifically, we divide the range of depth values into 40 logarithmic bins, accounting for the fact that the estimation error for larger depth values is naturally larger. During training, we learn a classifier to assign the pixels to the bins. During evaluation, we use a soft-weighted-sum inference: Every bin is represented by its mean depth in log space. A weighted sum of bins (weight = prediction score) is used as the final prediction.

For surface normal estimation, we use the triangular coding technique of~\cite{zeisl2014discriminatively}. First, a codebook is learned with k-means. The codewords form a Delaunay triangulation cover on the unit sphere. Any surface normal can thus be expressed as a weighted combination of the three codewords marking its triangle. During training, we learn a classifier to predict those codeword weights. Following~\cite{zeisl2014discriminatively}, we choose 40 codewords ($\widehat{=}$ 40 classes). Evaluation consists of two steps: (1) Find the triangle with maximum total probability. (2) Use the probabilities of the three codewords of that triangle to reconstruct the surface normal.

\begin{figure}[t]
    \begin{minipage}[c]{0.48\linewidth}
        \resizebox{\linewidth}{!}{%
            \begin{tikzpicture}

\definecolor{color0}{rgb}{0.4,0.76078431372549,0.647058823529412}
\definecolor{color1}{rgb}{0.988235294117647,0.552941176470588,0.384313725490196}

\pgfplotsset{
    scale only axis,
    height=6cm,
    width=3cm,
}

\begin{axis}[
    ybar,
    bar width=7pt,
    tick align=outside,
    tick pos=left,
    xticklabel style={rotate=40.0},
    xtick = {1,2},
    xticklabels={Depth,Normal},
    axis y line*=left, 
    xmin=0.5, xmax=2.5,
    x grid style=dashed,
    xminorgrids,
    ylabel={RMSE},
    ymin=0, ymax=0.85,
    major x tick style = {opacity=0},
    minor x tick num = 1,
    minor tick length=2ex,
    legend cell align={left},
	legend style={column sep=2pt},
	% legend pos=south east
]
\addplot[draw=black,fill=color1,error bars/.cd, y dir=both, y explicit, error mark=none]
	coordinates {
	(1,0.6104)  += (0,0.0041) -=(0,0.0041)
	};
\addplot[draw=black,fill=color0,error bars/.cd, y dir=both, y explicit, error mark=none]
	coordinates {
	(1,0.6159)  += (0,0.0013) -=(0,0.0013)
	};
\legend{Regression,Classification}
\end{axis}

\begin{axis}[
    ybar,
    bar width=7pt,
    tick align=outside,
    % tick pos=left,
    xticklabel style={rotate=40.0},
    xtick = {1,2},
    xticklabels={Depth,Normal},
    axis y line*=right,
    axis x line=none,
    xmin=0.5, xmax=2.5,
    x grid style=dashed,
    xminorgrids,
    ylabel={Mean angular error [\degree]},
    ymin=0, ymax=31,
    major x tick style = {opacity=0},
    minor x tick num = 1,
    minor tick length=2ex,
]
\addplot[draw=black,fill=color1,error bars/.cd, y dir=both, y explicit, error mark=none]
	coordinates {
	(2,20.94)  += (0,0.08) -=(0,0.08)
	};
\addplot[draw=black,fill=color0,error bars/.cd, y dir=both, y explicit, error mark=none]
	coordinates {
	(2,23.00)  += (0,0.09) -=(0,0.09)
	};
\end{axis}

\end{tikzpicture}
        }%
    \end{minipage}\hfill
    \begin{minipage}[c]{0.49\linewidth}
        \caption{Performance comparison of single task depth and surface normal estimation models, using either a regression or classification framework. Their similar performance confirms that we can exploit the classification scheme to form high-quality label regions for the label context.}
        \label{fig:class}
    \end{minipage}
\end{figure}
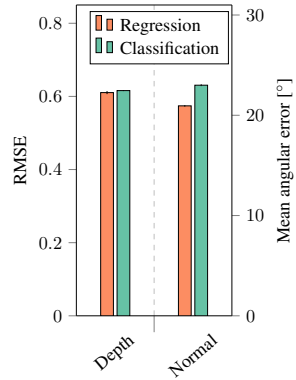

\begin{figure}[t]
    \includegraphics[width=\linewidth]{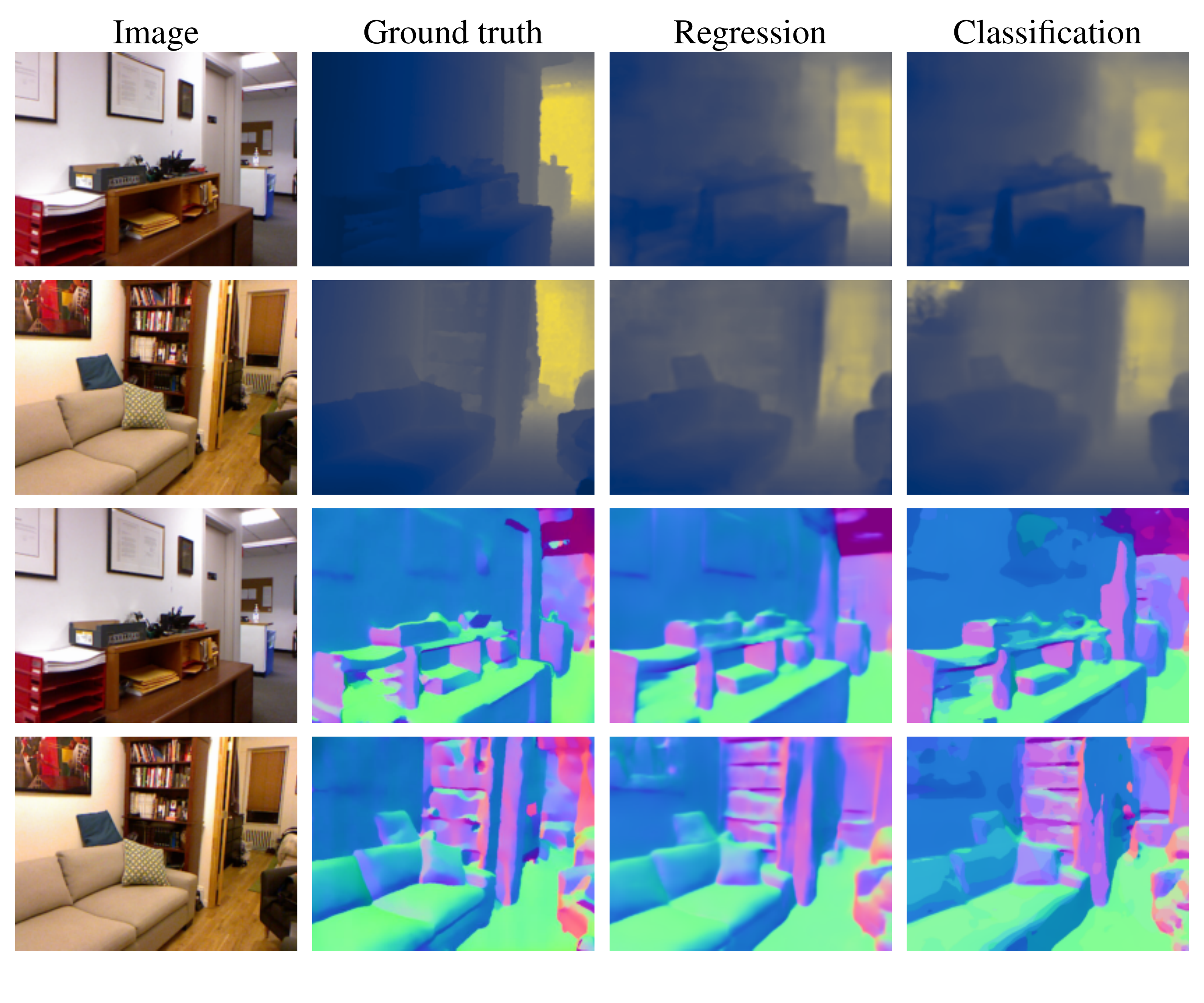}
    \caption{Qualitative NYUD-v2 comparison of regression and classification schemes for depth (top two rows) and surface normal (bottom two rows) estimation. Classification achieves satisfactory results on both tasks.}
    \label{fig:regrclass}
\end{figure}

To verify the above discretization schemes, we trained single task models accordingly, and compare them to the regression models in Fig.~\ref{fig:class}. The figure shows that the performance of classification---while slightly worse than regression---is satisfactory for both depth and surface normal estimation. The same conclusion can be drawn from a qualitative comparison, shown in Fig.~\ref{fig:regrclass}.

\beginappendixe
\section{Label Context: Performance Upper Bound}

To estimate the potential of label context for multi-modal distillation, we conduct experiments using ground truth label regions. Instead of predicting the spatial maps $A_n$ from the input image (see Sec.~\ref{subsec:similarity_context}), we directly use the ground truth data $A_n^{(GT)}$ to partition the label space into distinct regions. This provides an upper bound for the performance of label context distillation. Table~\ref{tab:bound} shows the results for both $T$-label and $S$-label context: The performance increases greatly (with the exception of $T$-label context for boundary detection), confirming that label region grouping is highly effective for multi-modal distillation.

\begin{table}
\begin{center}
\resizebox{\linewidth}{!}{%
\ra{1.3}
\Large
\begin{tabularx}{\textwidth}{@{}lYYYYr@{}}\toprule
Model & SemSeg~$\uparrow$ & Depth~$\downarrow$ & Normal~$\downarrow$ & Bound~$\uparrow$ & $\Delta_m$~[\%]~$\uparrow$ \\ \midrule
Single task & 38.02 & 0.6104 & 20.94 & 76.22 & 0.00 \\
\hdashline\noalign{\vskip 0.5ex}
$T$-label, GT & 46.71 & 0.5202 & 18.16 & 76.06 & 12.67 \\
$S$-label, GT & 47.71 & 0.5160 & 17.87 & 78.18 & 14.55 \\
\bottomrule
\end{tabularx}%
}
\end{center}
\caption{NYUD-v2 comparison of the performance upper bound of $T$-label and $S$-label context, using ground truth (GT) spatial region maps $A_n^{(GT)}$ (see Sec.~\ref{subsec:similarity_context}).}
\label{tab:bound}
\end{table}

\beginappendixf
\section{Context Type Search Reliability}
\label{sec:reliability}

We consider context type selection during architecture search as a rater decision. Since we repeat each run five times, we can evaluate the intra-rater reliability: The agreement among the five selected context types in all CP blocks. 

The most intuitive way to quantify agreement is through percentage agreement (\ie, counting the fraction of times a pair of runs agree on a decision). However, this measure does not take into account that agreement may happen purely due to chance. We thus report also Light's kappa~\cite{light1971measures}, which is an agreement score calculated by averaging Cohen's kappa~\cite{cohen1960coefficient} over all pairs of runs. Kappa statistics are corrected for chance agreement, with the drawback that their interpretation is less intuitive. A value of 0 indicates no agreement, \hbox{-1} indicates perfect disagreement, and 1 perfect agreement. We obtain an overall percentage agreement of 71.2\% and a Light's kappa of 0.48 for the search on NYUD-v2 with a HRNet18 backbone. 

We emphasize that reliability is not strictly necessary for an effective search algorithm. If there is no dominating choice of context type (\eg, none of the options lead to significant performance gain), then even a valid search algorithm is expected to be unreliable. In such cases, introducing a tie-breaking auxiliary objective could help promote convergence (\eg, a resource loss).

\beginappendixg
\section{How Important is Self-Attention in ATRC?}

The permutation testing results of Sec.~\ref{subsec:importance} can be utilized to partly address this question. We conclude there that self-attention constitutes the most important distillation module for 3 out of 4 investigated tasks. However, other cross-task connections contribute significantly too. To investigate further, we provide in Table~\ref{tab:extra_table} the performance of ATRC without self-attention. The multi-task performance $\Delta_m$ for this model is 0.87\% (\vs 1.56\% for the full ATRC), outperforming the single task configuration. This confirms that, even though self-attention is vital according to permutation testing, the cross-task distillation modules are able to provide a substantial performance boost on their own.

\begin{table}[h]
\begin{center}
\resizebox{\linewidth}{!}{%
\ra{1.3}
\begin{tabular}{@{}lcccccccccccr@{}}\toprule
\multirow{2}{*}{Distillation module} & \multicolumn{2}{c}{SemSeg~$\uparrow$} & & \multicolumn{2}{c}{Depth~$\downarrow$} & & \multicolumn{2}{c}{Normal~$\downarrow$} & & \multicolumn{2}{c}{Bound~$\uparrow$} & \multirow{2}{*}{$\Delta_m$~[\%]~$\uparrow$} \\
\cmidrule{2-3} \cmidrule{5-6} \cmidrule{8-9} \cmidrule{11-12} & mean & std. && mean & std. && mean & std. && mean & std.\\ \midrule
None (single task baseline) & 38.02 & 0.14 && 0.6104 & 0.0041 && 20.94 & 0.08 && 76.22 & 0.07 & 0.00 \\
None (multi-task baseline) & 36.35 & 0.26 && 0.6284 & 0.0034 && 21.02 & 0.06 && 76.36 & 0.05 & -1.89 \\ \hdashline\noalign{\vskip 0.5ex}
ATRC (ours) & 38.90 & 0.43 && 0.6010 & 0.0046 && 20.48 & 0.02 && 76.34 & 0.12 & \textbf{1.56} \\
ATRC (no self-attention) & 38.19 & 0.46 && 0.6032 & 0.0038 && 20.55 & 0.05 && 76.22 & 0.10 & 0.87 \\
\bottomrule
\end{tabular}
}
\end{center}
\caption{Effect of removing the self-attention blocks on NYUD-v2 with a HRNet18 backbone. First three lines correspond to the numbers reported in Table~\ref{tab:control}.}
\label{tab:extra_table}
\end{table}

\end{appendices}

\end{document}